\documentclass[titlepagetrue, pdflatex, sn-vancouver-ay]{sn-jnl}

\usepackage{graphicx}%
\usepackage{multirow}%
\usepackage{amsmath,amssymb,amsfonts}%
\usepackage{amsthm}%
\usepackage{mathrsfs}%
\usepackage{bbm}
\usepackage{hyperref}
\usepackage[title]{appendix}%
\usepackage{xcolor}%
\usepackage{textcomp}%
\usepackage{manyfoot}%
\usepackage{booktabs}%
\usepackage{algorithm}%
\usepackage{algorithmicx}%
\usepackage{algpseudocode}%
\usepackage{listings}%
\usepackage{tabularx}
\usepackage{soul}
\usepackage{xcolor}

\theoremstyle{thmstyleone}%
\theoremstyle{thmstyletwo}%

\theoremstyle{thmstylethree}%

\newcommand{\highlight}[1]{\sethlcolor{yellow}\hl{#1}}
\renewcommand{\highlight}[1]{#1}

\begin{document}

\title[MARS: A neurosymbolic approach for interpretable drug discovery]{MARS: A neurosymbolic approach for interpretable drug discovery}


\author*[1,2,3]{\fnm{Lauren Nicole} \sur{DeLong}}\email{L.DeLong@crukscotlandinstitute.ac.uk}

\author[2]{\fnm{Yojana} \sur{Gadiya}}\email{yojana.gadiya@enveda.com}

\author[1,3]{\fnm{Paola} \sur{Galdi}}\email{P.Galdi@crukscotlandinstitute.ac.uk}

\author[1]{\fnm{Jacques D.} \sur{Fleuriot}}\email{jacques.fleuriot@ed.ac.uk}

\author[2]{\fnm{Daniel} \sur{Domingo-Fern\'{a}ndez}}\email{daniel.domingo-fernandez@enveda.com}

\affil*[1]{\orgdiv{School of Informatics}, \orgname{University of Edinburgh}, \orgaddress{\street{10 Crichton St}, \city{Edinburgh}, \postcode{EH8 9AB}, \country{United Kingdom}}}

\affil[2]{\orgname{Enveda Therapeutics}, \orgaddress{\street{5700 Flatiron Parkway}, \city{Boulder}, \postcode{80301}, \state{Colorado}, \country{United States of America}}}

\affil[3]{\orgname{Cancer Research UK Scotland Institute}, \orgaddress{\street{Switchback Road}, \city{Glasgow}, \postcode{G61 1BD}, \country{United Kingdom}}}


\abstract{
\textbf{Background:} Neurosymbolic (NeSy) artificial intelligence describes the combination of logic or rule-based techniques with neural networks. Compared to neural approaches, NeSy methods often possess enhanced interpretability, which is particularly promising for biomedical applications like drug discovery. However, no clear guidelines exist to assess the biological plausibility of model interpretations.

\textbf{Methods:} To assess interpretability in the context of drug discovery, we devise a novel prediction task, called drug mechanism-of-action (MoA) deconvolution, with an associated, tailored knowledge graph (KG), \textit{MoA-net}. We then develop the \textit{MoA Retrieval System (MARS)}, a NeSy approach for drug discovery which leverages logical rules with \textit{learned} rule weights.

\textbf{Results:} Using MARS' interpretable features alongside domain knowledge, we find that MARS and other NeSy approaches on KGs are susceptible to reasoning shortcuts, in which the prediction of true labels is driven by ``degree-bias'' rather than the domain-based rules. Subsequently, we demonstrate ways to identify and mitigate this. Thereafter, MARS achieves performance on par with current state-of-the-art models while producing model interpretations aligned with known MoAs.

\textbf{Conclusion:} Through MARS, we showcase the novel task of computational MoA deconvolution. Our results emphasize the importance of using interpretable models, like NeSy ones, for applications in drug discovery. Specifically, by identifying and mitigating reasoning shortcuts, MARS MoA predictions which are biologically meaningful and, therefore, more reliable for downstream drug discovery research.}




\maketitle

\section{Background}\label{sec:intro}

Drug discovery (DD), the search for novel drugs to treat ailments, often involves screening thousands of small chemical compounds \citep{drug_screening}. Many computational approaches have been developed to accelerate and streamline this screening process \citep{gottlieb2011predict, gan2023drugrep}. Specifically, hundreds of such approaches operate upon knowledge graphs (KGs), in which nodes representing drugs, proteins, or medical conditions are connected by edges, representing the relationships between them \citep{survey_kg_metrics}. Typically, DD is formulated on a KG as a link prediction task between drugs and the corresponding medical conditions (indications) to be treated \citep{schultz2021drscai, rivas2022kgem}.

\begin{figure}[ht]
\begin{center}
\includegraphics[width=0.9\linewidth]{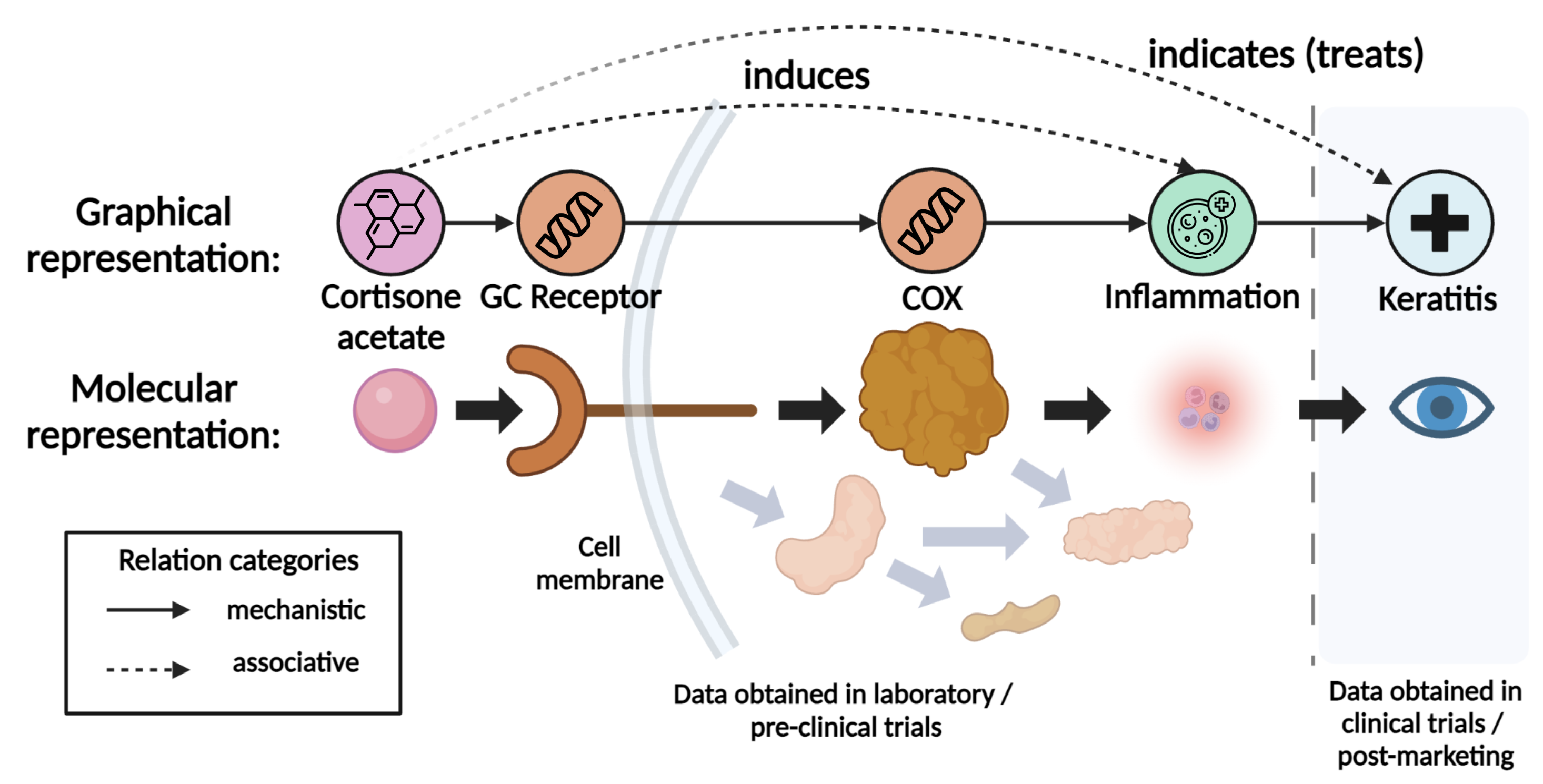}
\end{center}
\caption{MoA of cortisone acetate. Cortisone acetate upregulates the activity of the glucocorticoid (GC) receptor protein, which, in turn, downregulates the cyclooxygenase (COX) protein. Since COX is directly involved in creating inflammation, its inhibition reduces inflammation, thereby treating keratitis \mbox{\citep{drugmechdb}}. Data regarding protein interactions and biological processes (left) can be collected in a laboratory setting, whereas physiological effects like indications (right) are obtained during or after clinical trials. Created in https://BioRender.com.}
\label{fig:moa}
\end{figure}

In addition to predicting drug indications, it is also important to understand a drug's mechanism-of-action (MoA), the molecular processes by which it achieves its medicinal effect. As depicted in Fig.~\ref{fig:moa}, MoAs typically involve chains or paths of physical, molecular interactions induced by a drug \citep{crino2016mtor}. Revealing a drug's MoA, which we call \textit{MoA deconvolution}, informs researchers as to how each drug works, de-risks potential side effects \citep{palve2021offtargetdr, green2023offtargetkinase}, and helps to predict clinical trial success \citep{clin_trials2}. Unfortunately, traditional approaches for MoA deconvolution involve additional laboratory assays, which do not scale well to the magnitudes at which DD screening takes place \citep{lab_moa, drug_screening}. Our approach, called the MoA Retrieval System (MARS) \citep{MARS_code}, \highlight{could serve as a unique, AI-based alternative.}

 Specifically, MARS is a neurosymbolic (NeSy) approach. NeSy approaches combine logical rules with neural networks \mbox{\citep{lauren_neuroai}}, so they tend to possess enhanced \textit{interpretability} compared to state-of-the-art techniques on KGs \citep{gnn_survey}. Model interpretability involves transparency into the processes or patterns which led to certain predictions \citep{molnar2022}. Therefore, NeSy approaches have already been positioned as a promising avenue for MoA deconvolution \citep{liu2021neural, minerva}. However, interpretability is broadly defined \citep{molnar2022}, which poses an additional challenge: there are no clear guidelines for assessing the plausibility of model interpretations, especially for this novel task.

Although some previous studies present explainable or interpretable pipelines \highlight{\mbox{\citep{filtering_kg, liu2021neural, bioteque}}}, the corresponding explanations leverage \textit{associative} patterns: two nodes with mutual connections are likely to share other connections \citep{GBA_example}. For example, such methods utilize associations regarding a drug's pharmacological class \citep{filtering_kg}, side effects \citep{liu2021neural}, or known indications \citep{bioteque}. These \textit{associative} patterns, which are discovered during or after clinical trials, might be useful for repurposing clinically approved drugs \citep{schultz2021drscai, rivas2022kgem}. However, associative patterns are rare or absent for \textit{novel} compounds which have not yet undergone clinical trials. Furthermore, such patterns can not represent the MoA of a drug; instead, an MoA involves \textit{mechanistic} patterns. For example, \mbox{Figure~\ref{fig:moa}} shows the MoA of cortisone acetate, which involves physical, molecular interactions \citep{drugmechdb}. Therefore, we define \textit{computational MoA deconvolution} as the prediction of \textit{mechanistic} paths between drugs and their biological effects, such as biological processes (BPs). \highlight{BPs are a class of the Gene Ontology (GO) \mbox{\citep{gene2019gene}} representing the outcomes of molecular interactions.}

 To benchmark this novel task, we generate a tailored KG, called \textit{MoA-net} \citep{MoA_net_code}, from real-world, experimental data. Unlike existing KGs that are designed for DR \highlight{\mbox{\citep{hetionet, openbiolink}}}, the entities and relations in \textit{MoA-net} exclusively comprise \textit{molecular} interactions which could be captured in a typical laboratory setting. To perform MoA deconvolution, MARS is used upon \textit{MoA-net} to predict the links between nodes representing drugs and the BPs they cause. 

Several interpretable features of MARS provide insight into the drug's MoA, represented as a path between the two nodes. However, MARS' interpretable features also reveal a \textit{reasoning shortcut}. \highlight{A \textit{reasoning shortcut} describes a situation in which predictions are based on unintended semantics \mbox{\citep{reasoning_shortcuts, reasoning_shortcuts2}}. For example, take the classic MNIST addition task, popularly used to assess NeSy methods \mbox{\citep{manhaeve2018deepproblog}}. To accomplish the MNIST addition task, a model is trained to determine the sum of two handwritten digits. In some cases, models were found to predict the correct sum, despite having misclassified the digits \mbox{\citep{reasoning_shortcuts1}}. In \textit{this} study, we found that predictions upon \textit{MoA-net} are often driven by \textit{degree-bias}, rather than the rules representing domain knowledge. \textit{Degree-bias} occurs when a KG's \textit{node degree distribution}, describing the number of edges connected to each node, varies greatly \mbox{\citep{degree_bias_xswap}}. Consequently, this affects how accessible, discoverable, and influential certain nodes might be to various statistical and AI-based methodologies \mbox{\citep{degree_bias_ev1, degree_bias_ev2}}.}

To address this reasoning shortcut, we consider the desiderata from Marconato \textit{et al.} \citep{marconato2024bears} for making NeSy systems \textit{shortcut-aware}: (1) \textit{calibration}, high accuracy on concepts unaffected by reasoning shortcuts, (2) \textit{performance}, high accuracy despite reasoning shortcuts being present, and (3) \textit{cost effectiveness} achieved through simple mitigation strategies. Using these desiderata as guidelines, we make MARS shortcut-aware for more insightful predictions involving DD and MoA deconvolution. Ultimately, our study underscores the importance of evaluating the capabilities of NeSy models within applied domains: by evaluating model interpretations against specific domain knowledge, we can more easily identify and mitigate shortcuts.

\subsection{Related Work}\label{sec:related_work}

\subsubsection{Interpretable methods for drug discovery}

\highlight{Several previous methods for interpretable, computational DD lay the groundwork for MARS. Specifically, several of such methods use \textit{metapaths}. \textit{Metapaths} are abstract representations of paths in a KG \mbox{\citep{hetionet, noori2023metapaths}}, and they are particularly useful for capturing long-range dependencies between nodes which are several hops away from one another. \textit{Project Rephetio} \mbox{\citep{hetionet}}, in particular, was foundational for the use of metapaths in DD. The authors of Project Rephetio prioritized drug-disease associations by evaluating the prevalence of specific metapath patterns across a large, heterogeneous KG. To do so, they used degree-weighted path counts (DWPC) \mbox{\citep{himmelstein2015heterogeneous}}. In short, DWPCs enumerate all possible paths between source and target nodes, taking node degree into account. From this information, they predicted new drug-disease relations, and the metapaths used to make such predictions served as model interpretations. Another study by Kawichai \textit{et al.} \mbox{\citep{metapaths_go}} used a similar approach, but they extracted metapaths involving relevant GO entities, like BPs (see \mbox{Section~\ref{sec:intro}}). Their approach could effectively predict novel drug-disease associations to inform research on drug repurposing. Becuase metapaths were shown to be both useful and interpretable tools for this task, other studies applied them in different ways. For example, Ratajczak \textit{et al.} \mbox{\citep{filtering_kg}} leveraged metapaths from Project Rephetio, but they used them, instead, to filter the KG down to a more informative subgraph for drug repurposing. Another method, called \textit{Bioteque} \mbox{\citep{bioteque}}, obtained representative embeddings of KG paths following each metapath. Like the aforementioned studies, they also found that their embeddings were useful for predicting novel drug-disease associations. Although our article discusses the use of mechanistic versus associative patterns used by KG-based methods, a variety of methods exist which capture different concepts. For example, the \textit{GFlowNet} \mbox{\citep{gflownet}} uses paths in a graph to represent potential synthesis routes, involving chemical reactions, between reagents and a resultant drug. For a wider scope of interpretable computational DD, we refer the interested reader to broader reviews on this topic \mbox{\citep{review1, review2}}.}

\subsubsection{NeSy AI for MoA deconvolution}

Several NeSy approaches involve logical rules reflecting path-like patterns in biomedical KGs \highlight{for DD-related link prediction tasks.} \highlight{For example, an approach by Tian \textit{et al.} \mbox{\citep{bcm_citation1}} used a combination of path patterns (similar to that in Section \mbox{\ref{subsec:metapaths}}) and graph neural networks (GNNs) to predict drug side effects in an interpretable manner.} Similarly, \highlight{to predict drug indications,} Sudhahar \textit{et al.} \citep{sudhahar2024experimentally} investigated \textit{evidence chains}, paths explaining associations between drugs and diseases. However, these explanations were derived separately from indication predictions, via an additional rule-mining model \citep{meilicke2019anytime}\highlight{, much like that of Renaux \textit{et al.} \mbox{\citep{bcm_citation2}}, who used rule mining methods to unveil disease-causing gene interactions.} Other approaches \citep{minerva, liu2021neural, drance2021neuro} accomplished \highlight{biomedical link prediction for a variety of DD-related tasks} through deep reinforcement learning (RL), in which a neural network contributes toward the optimization of a reward function \citep{nesy_RL}. \highlight{One prominent example of such an approach is \textit{PoLo} \mbox{\citep{liu2021neural}}, which is based on \textit{MINERVA} \mbox{\citep{minerva}}. MINERVA involves a deep RL agent which traverses paths in a KG, receiving a reward when the first and last nodes of its path are known to have some relation of interest. PoLo expanded upon this by including an additional reward when the agent's trajectories utilized a set of predefined, informative metapaths. Critically, they applied a static, literature-derived weight to each of the metapaths to quantify how important each one should be to the agent. Consequently, after training, predictions were expected to utilize and align, to some extent, with the most important metapaths. Another study by Dranc\'{e} \textit{et al.} \mbox{\citep{drance2021neuro}} expanded on this idea by deriving the metapaths and their corresponding weights through rule-mining. Although weights remained static, they were mined directly from the KG, rather than sourced from an external study. In contrast to these studies, MARS \textit{learns} weights for each metapath, so the resultant weights are a more direct measure of relative importance toward predictions.} Additionally, while all these studies found that metapaths representing associative patterns were most frequently used, our study focuses upon paths involving mechanistic, molecular relations. In this study, we also identify a major risk: the approach may neglect to utilize rules in favor of \textit{other} semantics for reward optimization. This results in \textit{reasoning shortcuts}.

\subsubsection{Trustworthy NeSy methods and reasoning shortcuts}

\highlight{Like the deep RL approaches described above,} many other NeSy approaches are designed to abide by rules and domain knowledge \citep{drance2021neuro, dash2021lprules}. \highlight{For example, a study called Biomedical KG refinement with Embedding and Rules, or \textit{BioGRER} \mbox{\citep{zhao2020biomedical}}, used domain knowledge on the COVID-19 disease to construct relevant rules about it. By alternating between rules and a representative KG embeddings, BioGRER predicted novel drug indications and other biologically relevant information on COVID-19. Another NeSy approach, called \textit{Walking RDF and OWL} \mbox{\citep{alshahrani2017neuro}}, presented a pipeline for predicting drug indications in which ontological rules helped to predict new relations and augment the KG. Subsequently, a GNN was used to refine the augmented KG. Unlike the aforementioned approaches which utilize metapaths, methods which utilize GNNs or KG embeddings often suffer from local receptive fields \mbox{\citep{pasa2023empowering}}, experiencing peak performances at two layers \mbox{\citep{noor2023determining, guo2019learning}}. Nonetheless, they behave similarly in that they use rules to guide or inform a neural network as well as to explain biomedical link predictions. Thus, the use of rules and domain knowledge in NeSy AI often} portray such approaches as more trustworthy than neural, black box ones \citep{trustworthy}. Recent studies, however, have found that NeSy approaches may suffer from reasoning shortcuts, in which a model predicts the correct outcome via unintended semantics \citep{reasoning_shortcuts, reasoning_shortcuts2}. While reasoning shortcuts are not exclusive to NeSy methods \citep{RS_multihop_QA, LLM_RS}, they may be more easily overlooked when such approaches are portrayed as trustworthy.

\section{Methods}

Here, we present the MoA Retrieval System (MARS)\footnote{https://github.com/laurendelong21/MARS} \citep{MARS_code} to perform MoA deconvolution. MARS improves PoLo \citep{liu2021neural} (Section~\ref{sec:related_work}) by introducing dynamic, \textit{learned} rule weights. This differs from previous approaches, where weights are static and pre-computed (\textit{e.g.,} mined or literature-derived) \citep{liu2021neural, drance2021neuro}. As discussed further, these learned weights also make MARS shortcut-aware. To test MARS, we designed a novel KG, called \textit{MoA-net}, comprising data that is particularly relevant to predicting drug MoAs for \textit{new} compounds, as opposed to approved ones.

\subsection{Datasets: \textit{MoA-net} and its variants}\label{sec:datasets}

We designed our KG, \textit{MoA-net}\footnote{https://github.com/laurendelong21/MoA-Net} \citep{MoA_net_code}, specifically for MoA deconvolution. \textit{MoA-net} consists of \textit{triples}, two nodes connected by an edge, representing relations between drugs, proteins, and BPs (Table~\ref{tab:kg_dist}). In this study, we depict triples in \textit{MoA-net} as binary predicates. For example, \(\texttt{interacts}(\textsl{Protein}, \textsl{Protein})\) states that two \(\textsl{Protein}\) nodes are connected via the \texttt{interacts} relation. Such protein-protein interactions (PPIs) are derived from several real-world datasets comprising experimental data, including Custom KG \citep{drug2ways} and OpenBioLink KG \citep{openbiolink}. In addition to PPIs, \textit{MoA-net} contains causal relations between drugs and proteins, such as \texttt{upregulates} and \texttt{downregulates}, also from Custom KG and OpenBioLink KG. \highlight{The \textsl{BP} nodes as well as the causal relations between proteins and BPs come from experimentally-derived and expert-curated GO annotations available in the UniProt database \mbox{\citep{uniprot2015uniprot}} (release 2023.05), a gold-standard collection of protein data. Such GO annotations are entered into UniProt by several groups of designated researchers, developers, and curators \mbox{\citep{gene2021gene}}, and they denote which GO terms, including BPs, pertain to each protein in UniProt.}

\highlight{While other biomedical KGs exist \mbox{\citep{hetionet, openbiolink}}}, \textit{MoA-net} is unique from other KGs because it comprises drug-BP triples. These are derived from publicly available functional and biochemical assays in ChEMBL (v33), an open access database of bioactive compounds \citep{gaulton2012chembl}. BPs are the fundamental processes for the regulation, defense, and communication of cells, including processes like \textit{signal transduction} or \textit{inflammation} \citep{gene2019gene}. Thus, \textit{MoA-net} is particularly useful for DD within the context of \textit{novel} compounds because it comprises entities and relations which could be discovered in a laboratory setting \textit{before} clinical trials. By representing a drug's therapeutic effect as a BP, framing DD within a pre-clinical context is possible.

Between the three node types, five unique edge types, or relations, are defined and shown in Table~\ref{tab:kg_dist}. Notably, the inverse relations of all edge types in \mbox{Table~\ref{tab:kg_dist}}, which run in the opposite direction of causality, are also included. Of the 1,622 drug-BP triples in \textit{MoA-net}, 48 also had \textit{known} MoAs in a database called \textit{DrugMechDB} \citep{drugmechdb}, a manually-curated compendium of known drug MoAs. \highlight{\textit{DrugMechDB} contains evidence-derived, \textit{mechanistic} paths, or MoAs, between drugs and their respective molecular and physiological effects. Specifically, like \textit{MoA-net} and other widely used biomedical KGs \mbox{\citep{hetionet, openbiolink}}, paths in \textit{DrugMechDB} involve biomedical entities sourced from gold-standard biomedical databases.}

\begin{table}[ht]
\centering
\begin{tabular}{@{}lr@{}}
\toprule
Node type                    & Count  \\ \midrule
\(\textsl{Drug}\)                & 300    \\
\(\textsl{Protein}\)             & 9,301  \\
\(\textsl{Biological\ Process\ (BP)}\) & 86 \\ \midrule
Edge type                    & Count  \\ \midrule
 \(\texttt{interacts}(\textsl{Protein}, \textsl{Protein})\)  & 86,786 \\
\(\texttt{participates}(\textsl{Protein}, \textsl{Biological\ Process\ (BP)})\)     & 4,325  \\
\(\texttt{downregulates}(\textsl{Drug}, \textsl{Protein})\)  & 2,205  \\
\(\texttt{upregulates}(\textsl{Drug}, \textsl{Protein})\)    & 1,631  \\
\(\texttt{induces}(\textsl{Drug}, \textsl{Biological\ Process\ (BP)})\)             & 1,622 
\end{tabular}
\caption{Node and edge type distributions and provenances in \textit{MoA-net}.}
\label{tab:kg_dist}
\end{table}

Finally, we created two variants of \textit{MoA-net}. To investigate reasoning shortcuts, we used the Zietz \textit{et al.} \citep{degree_bias_xswap} implementation of XSwap \citep{xswap_og}, which swaps edges in a KG without affecting the distribution of node degrees. We called the resultant KG \textit{MoA-net-permuted}. Additionally, MARS, described in Section~\ref{sec:mars}, includes an optional, automatic KG trimming step to sparsify the KG. In essence, this step reduces edges of each class to below a user-specified threshold by iteratively removing those between the highest-degree nodes. \highlight{To create a sparse variant of the \textit{MoA-net}, we set this threshold to 10,000 PPIs. This specific value was selected because it reduced the PPIs to \mbox{\(\sim50\%\)} of the total edges, striking a balance between mitigating the skewness of the node degree distribution and preserving the global connectivity of the KG. We refer to this trimmed subgraph as \textit{MoA-net-10k} (see \mbox{Section~\ref{subsec:id_mitigate_bias}}).}

\subsection{MARS: a NeSy approach for MoA deconvolution}\label{sec:mars}

\begin{figure}[ht]
\begin{center}
\includegraphics[width=\linewidth]{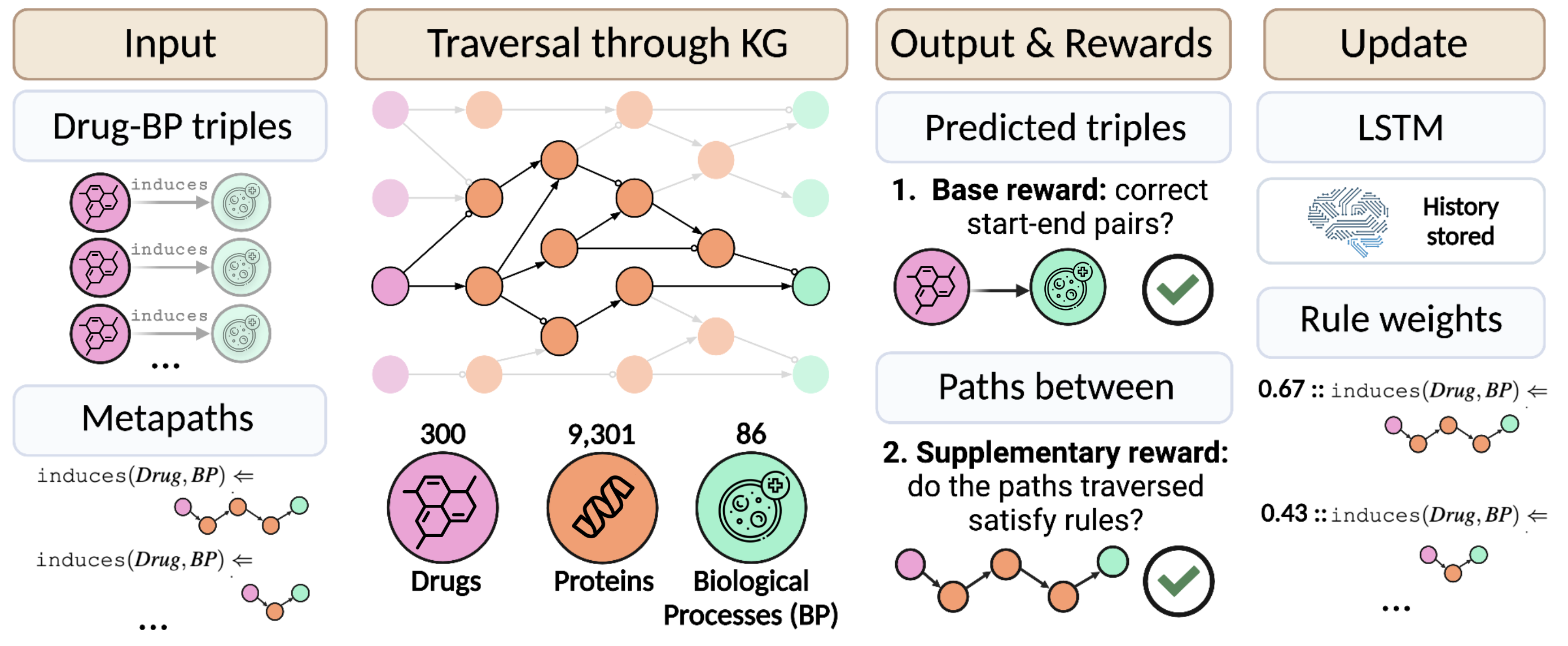}
\end{center}
\caption{Overview of the MoA retrieval system (MARS). Created in https://BioRender.com.}
\label{fig:mars}
\end{figure}

As shown in \mbox{Fig.~\ref{fig:mars}}, MARS takes two major inputs. The \textit{first} involves a KG. In this case, we test MARS on \textit{MoA-net} and its variants (see Section~\ref{sec:datasets}). Specifically, the input KG into MARS must contain triples involving some relation of interest. As shown in Figure~\ref{fig:moa}, some information about drug candidates, such their indications, is discovered \textit{during} or \textit{after} clinical trials, so this information is typically unavailable for novel compounds. Therefore, to understand each MoA as the \textit{biological} response to drug administration, we investigate MoAs as the relations between drugs and BPs (\textit{i.e.}, \(\texttt{induces}(\textsl{Drug}, \textsl{BP})\)). Specifically, we formulate this as a \textit{link prediction} task. In other words, we task MARS with determining whether edges of type \(\texttt{induces}\) exist between \(\textsl{Drug}\) and \(\textsl{BP}\) nodes in \textit{MoA-net}. Thereafter, new predictions regarding the \(\texttt{induces}\) relation serve as potential therapeutic outcomes for the chemical compound represented by the \(\textsl{Drug}\) node. As discussed in Section~\ref{sec:datasets}, this is a novel application in the KG field.

The \textit{second} input into MARS, as depicted in Figure~\ref{fig:mars}, includes metapath-based rules with corresponding weights.

\subsection{Metapath-based rules}\label{subsec:metapaths}

As mentioned in Section~\ref{sec:related_work}, metapaths are abstract representations of instantiated paths in a graph \citep{sun2011co, hetionet, noori2023metapaths}. For example, given the following path, \(P\), in \textit{MoA-net}: 

\begin{flalign}
\textsl{Cortisone acetate} &\xrightarrow{upregulates} \textsl{GC receptor} \notag \\
& \xrightarrow{interacts} \textsl{COX protein} \xrightarrow{participates} \textsl{Inflammation} &
\label{cortisone_acetate}
\end{flalign}

the corresponding metapath, \(\tilde{P}\), would be:
\begin{flalign}
    \textsl{Drug} \xrightarrow{upregulates} \textsl{Protein} \xrightarrow{interacts} \textsl{Protein} \xrightarrow{participates} \textsl{Biological Process (BP)}\label{cortisone_acetate_moa}
\end{flalign}

Within this study, metapaths can be understood as a sequence of triples within the KG structure, making them inherently interpretable. In MARS, metapaths are used as the bodies of logical rules, in which triples are connected by logical conjunctions (\(\land\)). Conjunctions indicate that, if all triples in the rule body are true, then the rule \textit{head} is evaluated as true. In other words, the rule is satisfied. The rule \textit{head}, the left side of the implication arrow (\(\Leftarrow\)), is a single triple representing the relation of interest between the first and last node types of the metapath, \textit{e.g.,}:
\begin{flalign}
\texttt{induces}(\textsl{Drug}, \textsl{BP}) &\Leftarrow \texttt{upregulates}(\textsl{Drug}, \textsl{Protein}_A) \land  \notag \\
&\hphantom{\Leftarrow} \texttt{interacts}(\textsl{Protein}_A, \textsl{Protein}_B) \land \texttt{participates}(\textsl{Protein}_B, \textsl{BP}) &
\label{example_rule}
\end{flalign}

For each metapath-based rule, \(M_i\), in a set of rules, \(\mathcal{M} = \{M_1, M_2, ..., M_m\}\), the rule weight is denoted by \(w(M_i)\in \mathbb{R}\), where \(0 \leq w(M_i) \leq 1\). Such a weight indicates the relative usefulness of the metapath-based rule to the prediction task. In Section~\ref{subsec:weight_updates}, we discuss how we initialize and compute these weights.

Using the \textit{hetnetpy} package \citep{hetnetpy}, we extracted all valid MoA patterns as metapaths from \textit{MoA-net} (see Section~\ref{sec:datasets}). We defined valid MoA patterns as metapaths comprising directed, \textit{mechanistic} paths between drug and BP nodes (see Table~\ref{tab:metapaths} in Appendix~\ref{app:metapaths}). \highlight{Based on MoAs found in \textit{DrugMechDB}, metapath length was limited to a maximum of four relations (or \textit{hops}).} We excluded metapaths depicting \textit{associative} patterns, such as those leveraging information about shared BP targets, from the set of metapath-based rules. \highlight{Specifically, associative patterns were identified based on paths in which a node was receiving directed, incoming edges from nodes of the same type.} Finally, each of these metapaths served as the body of a logical rule with \(\texttt{induces}(\textsl{Drug}, \textsl{BP})\) as the rule head, as in formula~\ref{example_rule}.

\subsection{Overview of MARS}\label{subsec:mars_overview}

Using a deep RL process, MARS trains an agent to take walks of length \(L\) through the KG to connect pairs of nodes having the pre-defined relation of interest. Here, that relation is \(
\texttt{induces}(\textsl{Drug}, \textsl{BP})\), which are masked from the agent during training. Each walk generates a path, \mbox{\(P\)}, such as the one in the previous section~\ref{subsec:metapaths}. This path, \mbox{\(P\)}, can also be understood as a series of \(L\) transitions: \(P := (e_{Drug} \xrightarrow{r_1} e_2 \xrightarrow{r_2} ... \xrightarrow{r_L} e_{L+1})\). With respect to classic RL terminology, each node can be considered a state, and each relation or transition between nodes can be considered an action. The agent may also remain at its current node. Ultimately, the goal of the agent is episodic: to find paths in which the starting node, \(e_{Drug}\) (the drug node), and the terminal node, \(e_{L+1}\), have the \texttt{induces} relation. By training the agent to do so, it can identify node pairs with the desired relationship, thus generalizing beyond the training set to predict novel pairs in a holdout, test set. In other words, while true positive predictions in the test set serve as validation, false positives are positioned as potentially \textit{novel} \(\texttt{induces}(\textsl{Drug}, \textsl{BP})\) predictions. 

Similarly to a Markov Decision Process \citep{MDP}, the agent moves based on its current position and its next possible actions, with no information about its target destination. Additionally, at each state or node in the agent's trajectory, the history of the agent’s \textit{previous} actions are encoded with a two-layer LSTM (see Appendix~\ref{app:hp_selection}). Given the current node, \(e_c\), its history would comprise entity and relation embeddings (\(\textbf{\textit{e}}\) and \textbf{\textit{r}}, respectively) of each entity and relation traversed to that point:

\begin{equation}
    \textbf{\textit{h}}_c = \text{LSTM}(\textbf{\textit{h}}_{c-1}, \textbf{\textit{r}}_{c-1}, \textbf{\textit{e}}_{c-1})
    \label{lstm}
\end{equation}

In this case, entity and relation embeddings can be understood as type-specific, learnable parameters of the LSTM. Using Eq.~\ref{lstm}, the agent's next action can be modeled as a probability distribution, \(\textbf{\textit{a}}_c\), where \(\textbf{\textit{W}}_1\) and \(\textbf{\textit{W}}_2\) are learnable weight matrices, and \(\textbf{\textit{A}}_c\) denotes the action space (adjacent nodes) from the current node:

\begin{equation}
    \textbf{\textit{a}}_c = softmax(\textbf{\textit{A}}_c(\textit{\textbf{W}}_2 \text{ReLU}(\textbf{\textit{W}}_1 [\textit{\textbf{h}}_c; \textbf{\textit{e}}_c])))
    \label{action_dist}
\end{equation}

Ultimately, MARS' policy, is to draw actions from a categorical probability distribution parametrized by Eq.~\ref{action_dist}. The parameters are trained to optimize the reward function, \(R(S_{L+1})\) (Eq.~\ref{eq:reward}), which is evaluated each time the agent completes \(L\) transitions from some starting node:

\begin{equation}
R(S_{L+1}) = \mathbbm{1}_{\{e_{L+1} = e_{BP}\}} + \mathbbm{1}_{\{e_{L+1} = e_{BP}\}}\lambda\sum_{i=1}^{m} w(M_i)  \mathbbm{1}_{\{\tilde{P} = M_i\}}\label{eq:reward}
\end{equation}

\begin{equation}
\mathbbm{1}_{\{A\}} = \begin{cases}
1 & \text{if } A = \text{true} \\
0 & \text{if } A = \text{false}
\end{cases}
\end{equation}

In short, the reward function, originally from Liu \textit{et al.} \citep{liu2021neural}, quantifies how successful \(P\) is according to two rewards (Figure~\ref{fig:mars}). The first, base reward indicates whether the terminal node in the path, \(e_{L+1}\), is one of the desired target (BP) nodes (\(e_{BP}\)) that forms a \highlight{known triple or \textit{true pair}} with the starting (drug) node, \(e_{Drug}\). Put simply: ``given an \mbox{\(\texttt{induces}(\textsl{Drug}, \textsl{BP})\)} triple, did the agent make a successful traversal between the drug and BP nodes?"

The second, supplementary reward, contingent upon the first, indicates whether the corresponding metapath, \(\tilde{P}\), matches any metapath-based rule, \(M_i\). The second reward is also proportional to the rule's corresponding weight, and MARS updates these weights during training. These updates are accomplished through a novel algorithm called \textit{two-hop joint probability}, or \(P_{2H}\) (Section~\ref{subsec:weight_updates}). Therefore, the agent is not only encouraged to find connections between true pairs of nodes, but it is also guided toward paths which resemble known MoAs. Thus, MARS has two key interpretable features: (1) paths between nodes which serve as potential MoA predictions, and (2) learned rule weights which serve as a proxy for the importance of each rule.

\subsection{MARS dynamically updates rule weights}\label{subsec:weight_updates}

After MARS is executed, its learned rule weights reflect each rule's relative usefulness in the prediction task. Rule weights are updated after every batch, which includes a subset of the training triples. This means that, during training, the weight updates drive the agent toward more informative paths and bypass the assumptions that pre-assigned rule weights are correct. This eliminates the need for pre-computed or literature-derived rule weights; thus, we initialize all weights uniformly as 0.5, a \textit{medium} level of importance. We tested \textit{two} different approaches for updating rule weights.

\subsubsection{Naive weight updates}\label{naive_updates}

The first method for updating rule weights is called \textit{naive updates}\highlight{, and it lays the groundwork for the \mbox{\(P_{2H}\)} updates, covered in the next section.} The \textit{naive} way to implement weight updates (\(\text{MARS}_{\text{naive}}\)) is to increase weights according to the frequency at which each metapath-based rule, \mbox{\(M_i\)} is satisfied. \highlight{In other words, frequency serves as a heuristic for how often the agent makes successful predictions with a given metapath.} Formally, this involves recording the observed frequency, \(O_{M_i}\), at which each metapath-based rule is satisfied in each batch. The observed frequency is then normalized by the batch-specific expected frequency, \(E\), which assumes that every rule has a uniform probability across the total number of occurrences in that batch. Therefore, \mbox{\(E\)} is the same for every metapath-based rule. Ultimately, this produces a metric, \(\mu_{M_i}\) (Eq.~\ref{eq:normalization}) in which \(\mu_{M_i} > 1\) indicates usefulness (the agent used that rule more than others), and \(\mu_{M_i} < 1\) indicates otherwise.

\begin{equation}
    \mu_{M_i}= O_{M_i} / E\label{eq:normalization}
\end{equation}

Then, using Eq.~\ref{eq:update}, denoted by \(\Phi\), and Eq.~\ref{eq:new_weight}, \(\mu_{M_i}\) is used to update the weight of a rule, \(w(M_i)\). 

\begin{equation}
    \Phi(\mu_{M_i}, w(M_i)) = w(M_i) \times 2\alpha(\frac{\mu_{M_i}-1}{\mu_{M_i}+1})\label{eq:update}
\end{equation}

\begin{equation}
    w(M_i)' = w(M_i) + \Phi(\mu_{M_i}, w(M_i))\label{eq:new_weight}
\end{equation}

\highlight{Several measures are taken to allow weight updates to happen gradually. Firstly, \mbox{Eq.~\ref{eq:update}} is regularized by the hyperparameter \mbox{\(\alpha \in \mathbb{R}\)}, where \mbox{\(0 \leq \alpha \leq 1\)}, to control how subtle or drastic the weight update is, respectively. If \mbox{\(\alpha = 0\)}, no weight updates are made. We selected \mbox{\(\alpha\)} via hyperparameter optimization (\mbox{Appendix~\ref{hyperparameters}}). Secondly, given that \mbox{\(\mu_{M_i} \geq 0\)}, the range of \mbox{Eq.~\ref{eq:update}} is \mbox{\([-w(M_i)2\alpha, w(M_i)2\alpha]\)}. Therefore, if \mbox{\(\alpha=1\)}, a weight can only, at most, triple during a single update. 
Thirdly, weight updates are executed after each batch, but batches only capture a fraction of possible trajectories through the KG. To increase the variety of metapaths seen, MARS' agent performs \textit{multiple} traversals, or \textit{rollouts}, for each triple in a batch. Despite this, however, certain metapaths may still go under- or over- sampled in a single batch, resulting in drastic weight fluctuations.

To avoid extreme values that may result from sampling biases, \mbox{\(\mu_{M_i}\)} is further restricted as described in \mbox{Eq.~\ref{eq:bound_mu}} by lower and upper bounds in \mbox{Eqs.~\ref{eq:lowerbound}} and \mbox{\ref{eq:upperbound}}. Specifically, \mbox{\({\text{batch size} \times \text{rollouts}}\)} describes the number of opportunities for which the agent could traverse a metapath, so the bounds are designed relative to this.  For naive updates, \mbox{\(\rho\)} is the total number of metapath-based rules.}

\begin{equation}
    \mu_{min}=\frac{\rho}{\text{batch size} \times \text{rollouts}} \label{eq:lowerbound}
\end{equation}

\begin{equation}
    \mu_{max}=\rho \times \text{batch size} \times \text{rollouts} \label{eq:upperbound}
\end{equation}

\begin{equation}
    \mu_{M_i} = \text{min}(\mu_{max}, \text{max}(\mu_{min}, \mu_{M_i})) \label{eq:bound_mu}
\end{equation}

\highlight{To understand the intuition behind \mbox{Eqs.~\ref{eq:bound_mu} and~\ref{eq:lowerbound}}, recall that \mbox{\(\mu_{M_i}\)} is a ratio to approximate the usefulness of a rule. If \mbox{\(\mu_{M_i}<1\)}, the corresponding weight, \mbox{\(w(M_i)\)}, will decrease, and if \mbox{\(\mu_{M_i}>1\)}, it will increase. If \mbox{\(\rho\)} is large, it will be more difficult for the agent to sample all metapath-based rules. Therefore, to reduce the impact of under-sampling on a weight update, \mbox{Eq.~\ref{eq:lowerbound}} defines a larger lower bound in such cases. If \mbox{\(\rho\)} is small, the agent is likely to encounter each metapath-based rule more frequently within a single batch. In this scenario, over-sampling is a risk, so \mbox{Eq.~\ref{eq:upperbound}} scales the upper bound constrains a positive weight update by the relative size of the rule space. Finally, if the agent finds zero occurrences of \textit{any} metapath-based rule, no weight updates are made at all.}

\subsubsection{\(P_{2H}\) weight updates}\label{p2h_updates}

The second, more complex method to update weights, which is also a novel component of this work, is coined \textit{two-hop joint probability} (\(P_{2H}\)). In essence, this metric approximates the usefulness of metapath-based rules based on full \textit{and} partial matches. Pseudocode for \(P_{2H}\) can be found in Algorithm 1 below. In summary, each \textit{two-hop fragment} is extracted from every path traversed by the agent (lines 2-8).

\begin{algorithm}
\caption{\(P_{2H}\) weight updates}
\begin{algorithmic}[1]
    \For{each batch, \(\beta\)}
        \State \(\mathcal{F} \gets\) [empty list]
        \For{each path, \(\mathcal{P}\), that the agent traverses,}
            \If{the agent found a true pair}
                \State \(\hat{P} \gets\) metapath(\(\mathcal{P}\)) \Comment{extract the metapath}
                \State \(\mathcal{F} \gets \mathcal{F} + \text{extract\_fragments}(\hat{P})\) \Comment{a list of two-hop fragments seen}
            \EndIf
            \EndFor
        \State \(E \gets 1 /\) num. unique fragments in \(\mathcal{F}\)
        \For{each unique fragment, \(f_j\), in \(\mathcal{F}\)}
            \State \(O_{f_j} \gets \text{count}({f_j})\)
            \Comment{\highlight{observed frequency of the fragment}}
        \EndFor
        \For{each metapath-based rule body, \(M_i\), in \(\mathcal{M}\)}
            \State \(\theta \gets \text{extract\_fragments}(M_i\)) \Comment{a list of the fragments in the metapath}
            \State \(P_{2H}(M_i) \gets \prod_{f=1}^{len(\theta)} \frac{O_{f_j}}{E}\)  \Comment{ratio of observed / expected frequency, as in Eq.~\ref{eq:normalization}}
            \State \(w(M_i) \gets w(M_i) + \Phi(P_{2H}(M_i), w(M_i))\) \Comment{use Eq.~\ref{eq:update} to adjust rule weight}
        \EndFor
    \EndFor
\end{algorithmic}
\end{algorithm}

Two-hop fragments are defined as the metapaths involving two consecutive relations in a given path or trajectory. For example, the two-hop fragments of the path expressed in \mbox{Eq.~\ref{cortisone_acetate}} would be:
\begin{itemize}
    \item \(\texttt{upregulates}(\textsl{Drug}, \textsl{Protein}) \land \texttt{interacts}(\textsl{Protein}, \textsl{Protein})\), as well as
    \item \(\texttt{interacts}(\textsl{Protein}, \textsl{Protein}) \land \texttt{participates}(\textsl{Protein}, \textsl{BP})\)
\end{itemize}

After all the two-hop fragments are extracted from the agent's trajectories in a given batch, weight updates can be computed similarly to naive updates (\mbox{Section~\ref{naive_updates}}). Here, \mbox{\(E\)} denotes the expected frequency of each fragment, computed as the reciprocal of the number unique fragments encountered (line 9). The observed frequencies are also adjusted in a similar manner, describing the number of times each two-hop fragment, \mbox{\(f_j \in \mathcal{F}\)}, is encountered (\mbox{\(O_{f_j}\)}) (lines 10-12). With these alterations, \mbox{Eq.~\ref{eq:normalization}} is used to compute \mbox{\(\mu_{f_j}\)}.

To compute weight updates, the two-hop fragments are also determined for each of the metapath-based rules (line 14). For example, take the following metapath-based rule, \mbox{\(M_{example}\)}:
\begin{flalign}
M_{example} &:= \texttt{induces}(\textsl{A}, \textsl{E}) \Leftarrow \texttt{upregulates}(\textsl{A}, \textsl{B}) \land \texttt{interacts}(\textsl{B}, \textsl{C}) \land \notag \\ 
&\hphantom{:= \texttt{induces}(\textsl{A}, \textsl{E}) \Leftarrow \texttt{upregulates}} \texttt{interacts}(\textsl{C}, \textsl{D}) \land \texttt{participates}(\textsl{D}, \textsl{E}) &
\label{example_metapath}
\end{flalign}
where two-hop fragments are pairs of binary predicates which share a variable:
\begin{itemize}
    \item \(\textsl{Fragment 1}: \texttt{upregulates}(\textsl{X}, \textsl{Y}) \land \texttt{interacts}(\textsl{Y}, \textsl{Z})\)
    \item \(\textsl{Fragment 2}: \texttt{interacts}(\textsl{X}, \textsl{Y}) \land \texttt{interacts}(\textsl{Y}, \textsl{Z})\)
    \item \(\textsl{Fragment 3}: \texttt{interacts}(\textsl{X}, \textsl{Y}) \land \texttt{participates}(\textsl{Y}, \textsl{Z})\)
\end{itemize}

Here, the probability of each metapath-based rule is computed as the joint probability of its fragments (line 15 of Algorithm 1). For example, the  \mbox{\(P_{2H}\)} metric for the metapath-based rule shown in \mbox{formula~\ref{example_metapath}} would be computed as \(P_{2H}(M_{example}) = p(\textsl{Fragment 1}) \times p(\textsl{Fragment 2}) \times p(\textsl{Fragment 3})\). Notably, there are two caveats. Firstly, to account for partial metapath matches, the definition of \textit{conjunction} here is relaxed, allowing truth to be evaluated on the \textit{fragment} level. Secondly, metapath fragments do \textit{not} necessarily represent independent events, but the way in which the \(P_{2H}\) metric is computed suggests otherwise. However, to avoid complex computation involving conditional probabilities for dependent events \citep{russell2010artificial}, independence is assumed, and the \(P_{2H}\) metric serves as an \textit{approximation} for the empirical probabilities of metapath-based rules. From this \mbox{\(P_{2H}\)} metric, weights can be updated using \mbox{Eqa.~\ref{eq:update} and~\ref{eq:new_weight}} from \mbox{Section~\ref{naive_updates}} (line 16 of Algorithm 1). Notably, for \(P_{2H}\) updates, \mbox{\(\rho\)} in Eqs.~\ref{eq:upperbound} and~\ref{eq:lowerbound} is the total number of unique two-hop fragments possible. Ultimately, MARS with \(P_{2H}\) updates (\(\text{MARS}_{P_{2H}}\)) uses \textit{all} information from successful trajectories.


We describe implementation details and hyperparameter selection in Appendices~\ref{app:implementation} and \ref{app:hp_selection}, respectively.

\subsection{Evaluation}\label{label:evaluation}

As mentioned in Section~\ref{sec:mars}, we test MARS upon \textit{MoA-net}, and performance is assessed based on link prediction over drug-BP triples. Therefore, we split the drug-BP triples within \textit{MoA-net} into training (60\%), validation (20\%), and test (20\%) sets. Models are evaluated using Hits@\(k\), where \(k \in \{1, 3, 10\}\) and mean reciprocal rank (MRR), optimizing for the latter. Hits@\(k\) reports the proportion of times the correct results are in the top \(k\) ranked entries, while MRR reports how highly ranked the first correct item is amongst ranked results \citep{survey_kg_metrics}. \highlight{In addition to these \textit{standard} metrics, we report \textit{pruned} metrics: these are computed \textit{solely} on a subset of the predictions that utilized one of the pre-defined metapath-based rules (see \mbox{Table~\ref{tab:metapaths}}), \textit{excluding} all predictions which did not satisfy a rule.} Notably, pruned metrics help us assess the \textit{calibration} desideratum, as introduced in Section~\ref{sec:intro}, since rule-based predictions follow the expected model semantics.

We conduct an extensive benchmark of MARS against five different baseline KGE models, two state-of-the-art NeSy RL methods, and one network measure. Based on previous biomedical KG benchmarks \citep{rivas2022kgem}, the KGE baseline models include ComplEx \citep{ComplEx}, RotatE \citep{RotatE}, MuRE \citep{MuRE}, CompGCN \citep{compgcn}, and PairRE \citep{pairre}. The two RL baseline models include PoLo \citep{liu2021neural} and its predecessor MINERVA \citep{minerva} (Section~\ref{sec:related_work}), which is not guided by rules. We describe hyperparameter selection for baseline models in Appendix~\ref{app:hp_selection}. Lastly, the network measure baseline includes prioritization of drug-BP triples based on DWPC (see Section~\ref{sec:related_work}). We train and evaluate all baseline models on the same data splits as MARS on \textit{MoA-net-10k}.

Finally, \(\text{MARS}_{P_{2H}}\) has two key interpretable features: firstly, all successful trajectories are recorded, serving as potential MoA predictions. This allows for a comparison between MARS' \textit{predicted} MoAs of 48 drug-BP pairs against their \textit{known} MoAs (see Section~\ref{sec:datasets}). Secondly, learned rule weights serve as a proxy for the importance of each metapath-based rule. This helps to determine whether agent trajectories are biased toward certain types of paths. Alongside the pruned metrics, these features help us evaluate MARS' alignment with domain knowledge.

\section{Results}\label{sec:results}

\highlight{Within the next sections, we present the results of \mbox{\(\text{MARS}_{P_{2H}}\)} and compare them against relevant baseline models. Specifically, we report standard and pruned metrics (see \mbox{Section~\ref{label:evaluation}}) to evaluate performance as well as Marconato \textit{et al.}'s \mbox{\citep{marconato2024bears}} shortcut-aware desiderata, introduced in \mbox{Section~\ref{sec:intro}}. In summary, we initially found that both variants of MARS, as well as PoLo, make predictions which, primarily, avoid using metapath-based rules. Instead, all three methods utilized paths following associative patterns. After several subsequent experiments, we verified that this reasoning shortcut was likely due to node degree bias in \textit{MoA-net}. Thereafter, we found that \mbox{\(\text{MARS}_{P_{2H}}\)} can mitigate this reasoning shortcut on a trimmed version of \textit{MoA-net}.}

\subsection{Associative patterns improve accuracy but offer limited practical use}\label{subsec:GBA}

\begin{figure}[h!]
    \centering
    {\includegraphics[width=0.5\textwidth]{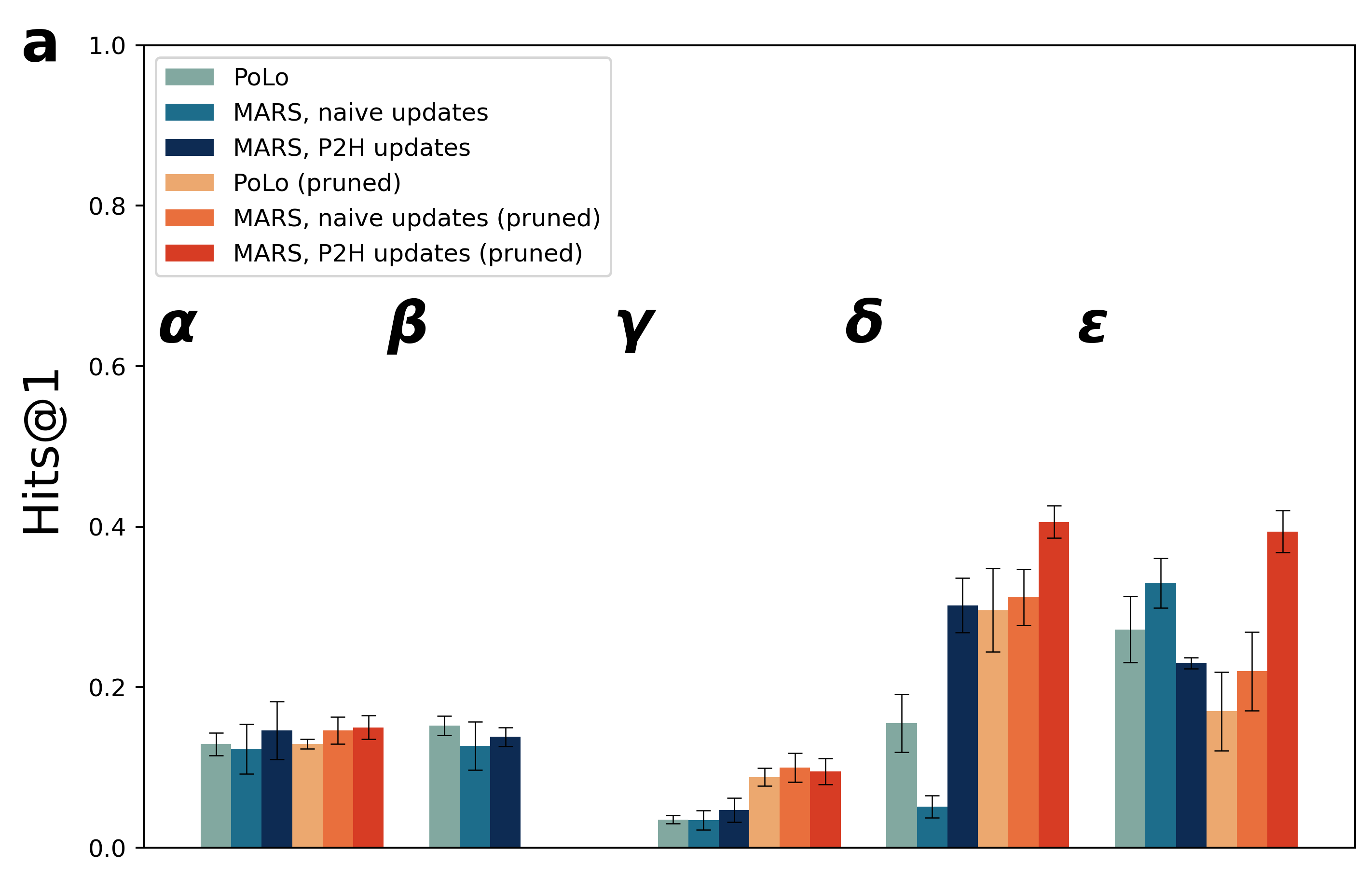}}{\includegraphics[width=0.5\textwidth]{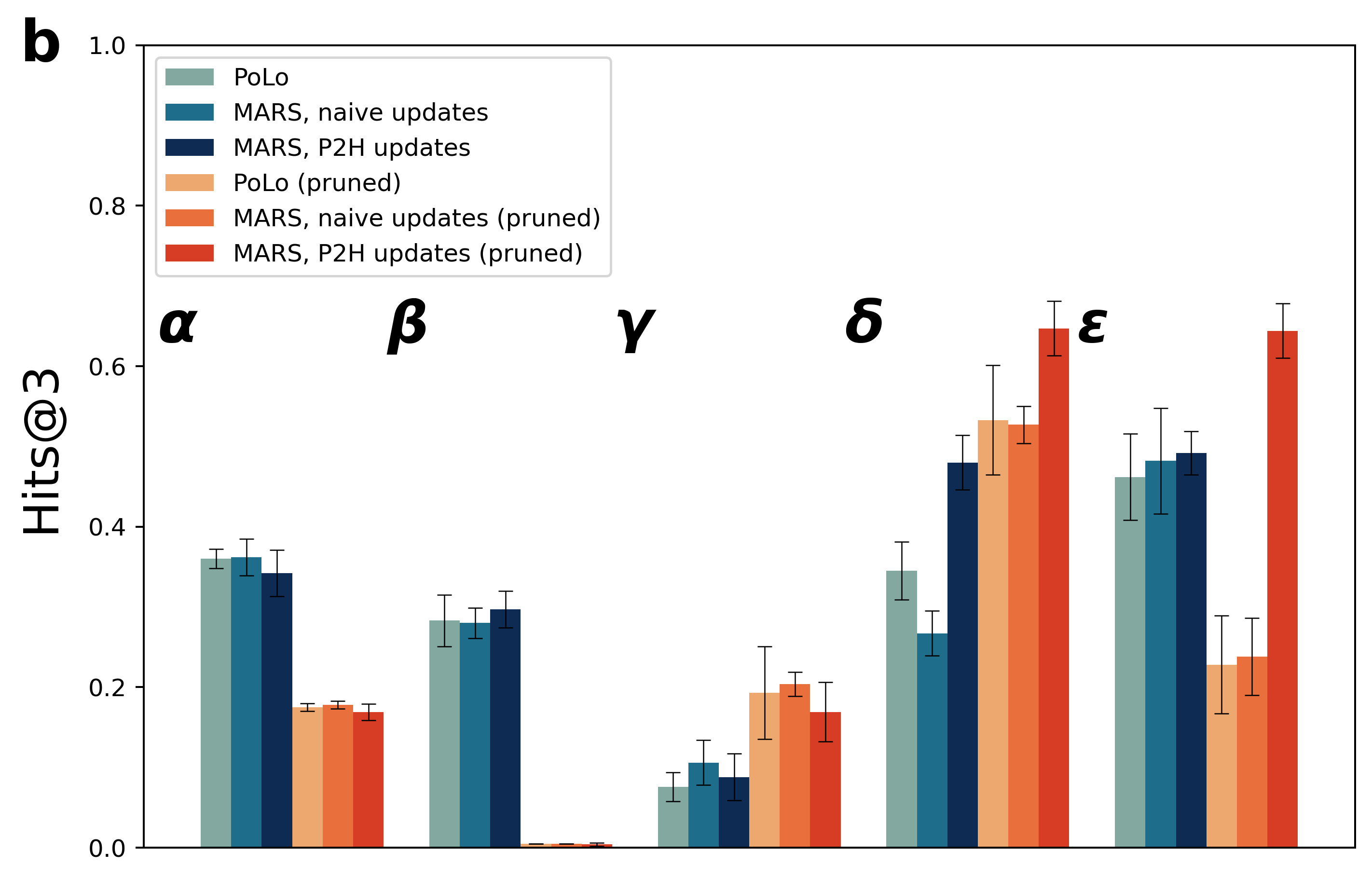}}
     {\includegraphics[width=0.5\textwidth]{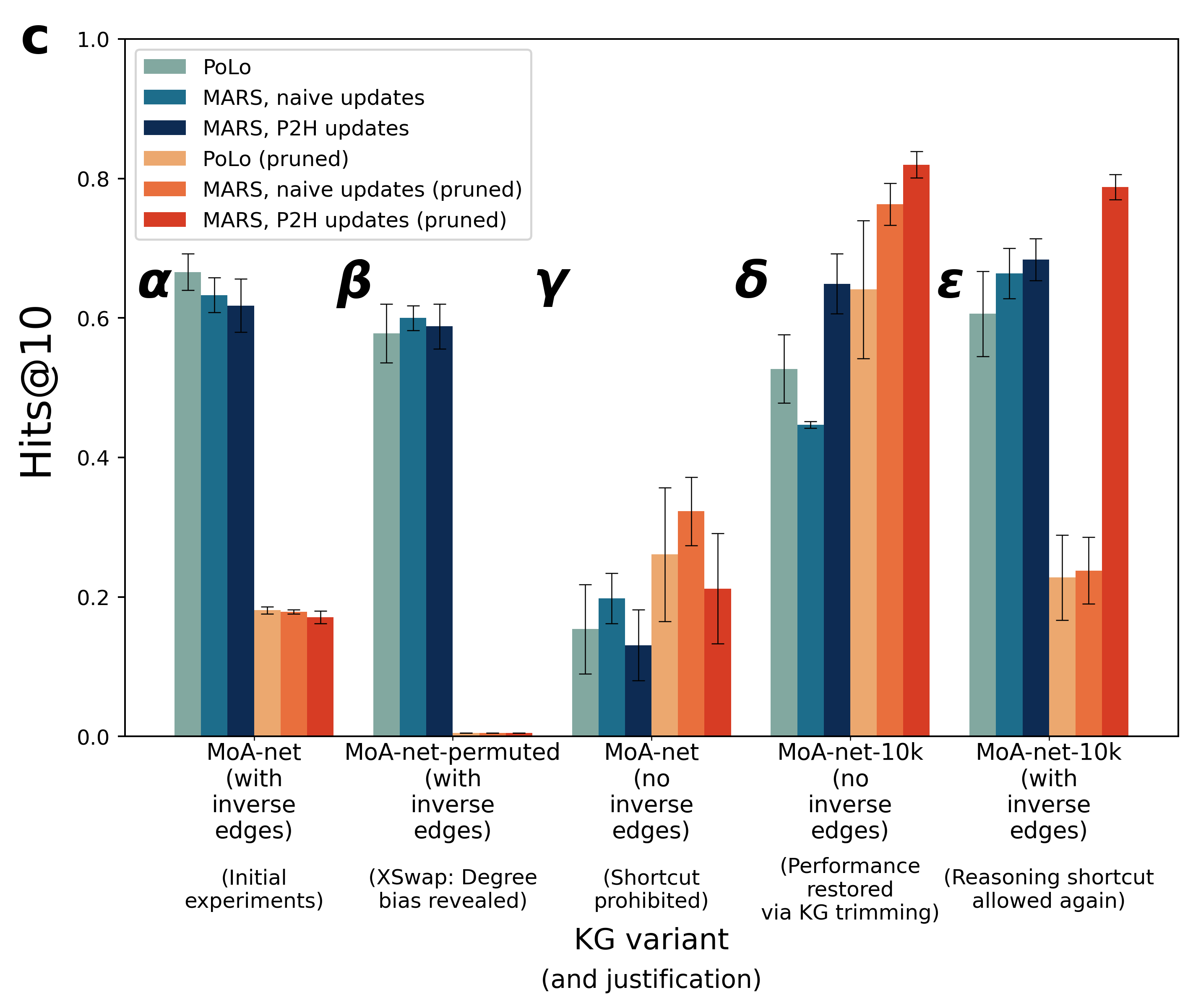}}{\includegraphics[width=0.5\textwidth]{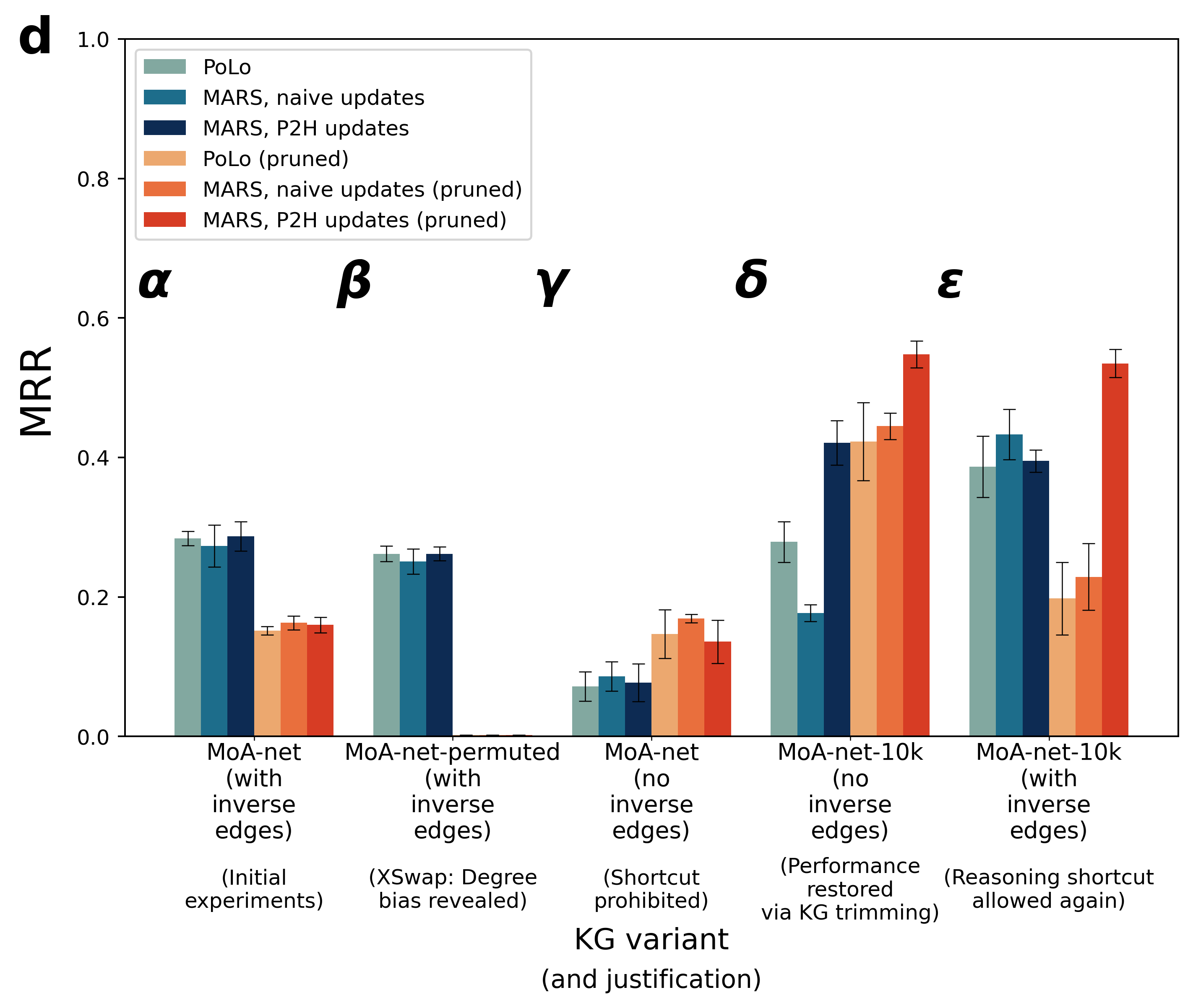}}
    \caption{\textbf{a-d}. Hits@\{1, 3, 10\} and MRR, respectively, for \(\text{MARS}_{P_{2H}}\) compared to PoLo and \(\text{MARS}_{\text{naive}}\) upon several variants of \textit{MoA-net}. Each bar is the average and standard deviation across five independent training and testing iterations. From left to right on each panel: the little change between the initial metrics upon \textit{MoA-net} (represented by \(\alpha\)) in comparison to the standard \textit{MoA-net-permuted} metrics (\(\beta\)) provides evidence that predictions are influenced by degree bias, resulting in a reasoning shortcut. Thereafter, inverse edges were removed to prohibit the reasoning shortcut, hindering performance (shown in \(\gamma\)). Performance was restored upon MoA-net-10k with the KG trimming step (\(\delta\)), with \(\text{MARS}_{P_{2H}}\) showing the best standard and pruned metrics. Finally, \(\text{MARS}_{P_{2H}}\) maintains high pruned metrics even when inverse edges (and reasoning shortcuts) are re-introduced (\(\epsilon\)).}\label{fig:metrics}
\end{figure}

After an initial set of experiments on \textit{MoA-net}, the results were already different than expected. Specifically, the pruned metrics were consistently lower than standard ones (Figure~\ref{fig:metrics}-\(\alpha\)), indicating that the metapath-based rules were \textit{not} being utilized in most predictions. This can happen because rule-based rewards are contingent upon a true pair being found (Eq.~\ref{eq:reward}). Additionally, amongst recorded trajectories, most did \textit{not} follow the metapath-based rules; instead, most trajectories used the following \textit{associative} pattern, involving inverse edges:
\begin{flalign}
\texttt{induces}(\textsl{Drug}_1, \textsl{BP}_2) &\Leftarrow \texttt{induces}(\textsl{Drug}_1, \textsl{BP}_1) \land \texttt{induces}(\textsl{Drug}_2, \textsl{BP}_1) \notag\\ 
&\hphantom{\Leftarrow \texttt{induces}(\textsl{Drug}_1, \textsl{BP}_1)} \land \texttt{induces}(\textsl{Drug}_2, \textsl{BP}_2) &
\label{mars_assoc_pattern}
\end{flalign}
This associative pattern indicates that two drugs inducing a common BP also likely induce another BP. This type of pattern was also present in the original PoLo study by Liu \textit{et al.} \citep{liu2021neural}, in which the most used pattern was the following:
\begin{flalign}
\texttt{treats}(\textsl{Drug}_1, \textsl{Disease}) &\Leftarrow \texttt{causes}(\textsl{Drug}_1, \textsl{Side Effect}) \land \texttt{causes}(\textsl{Drug}_2, \textsl{Side Effect}) \notag\\
&\hphantom{\Leftarrow \texttt{causes}(\textsl{Drug}_1, \textsl{Side Effect})} \land \texttt{treats}(\textsl{Drug}_2, \textsl{Disease}) &
\label{polo_assoc_pattern}
\end{flalign}

In Liu \textit{et al.}, the idea that \mbox{pattern~\ref{polo_assoc_pattern}} was commonly used by the agent had served as validation that the agent was, indeed, using the metapath-based rules. This is because \mbox{pattern~\ref{polo_assoc_pattern}} was one of the metapath-based rules which could be satisfied to achieve the supplementary reward (see \mbox{Section~\ref{subsec:mars_overview}}). Additionally, in its original study, PoLo had obtained high standard \textit{and} pruned metrics, thereby reinforcing this notion. However, high performance metrics do not guarantee biologically meaningful results. When the results from Liu \textit{et al.} \mbox{\citep{liu2021neural}} are reproduced, the same standard metrics can be achieved -- even when \mbox{pattern~\ref{polo_assoc_pattern}} is \textit{not} used to guide the agent (see \mbox{Appendix~\ref{app:polo}}). Additionally, \mbox{pattern~\ref{polo_assoc_pattern}} is \textit{still} the most commonly used pattern, even when it is not used as a metapath-based rule. This suggests that, although this associative rule may serve as a \textit{plausible} model explanation, it does not necessarily guide model training. From this, it was clear that neither PoLo's nor MARS' agents were using the supplementary reward in \mbox{Eq.~\ref{eq:reward}}. In the next section, MARS' interpretable features helped to understand why this was happening.

\subsection{\(P_{2H}\) updates reveal reasoning shortcuts via degree bias}\label{subsec:p2h_reveals_bias}

\begin{figure}[h!]
  \begin{center}
  \includegraphics[width=\linewidth]{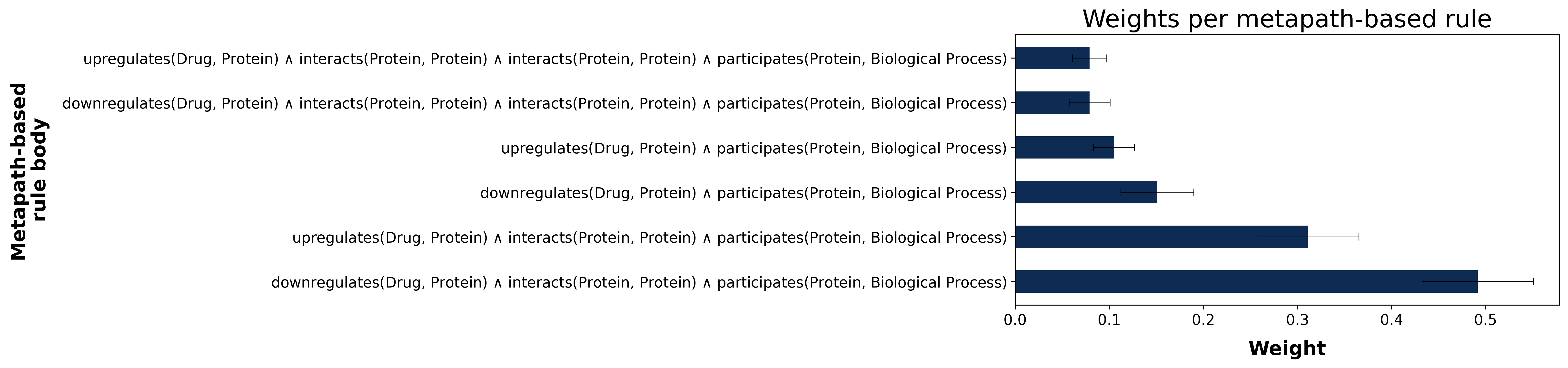}
  \end{center}
  \caption{Metapath-based rule weights from \(\text{MARS}_{P_{2H}}\) on \textit{MoA-net} (Figure~\ref{fig:metrics}-\(\alpha\)). Each bar is the average and standard error across five independent training and testing iterations. Paths involving consecutive PPIs (\(\texttt{interacts}(\textsl{Protein}, \textsl{Protein})\)), the most common relation type, have consistently lower weights.}
  \label{fig:confidences}
\end{figure}

In addition to analyzing the agent trajectories, we used the rule weights with \(P_{2H}\) updates to assess how informative each of the metapath-based rules was in making predictions. In particular, we found that \(\text{MARS}_{P_{2H}}\) weights for paths involving consecutive PPIs (see Section~\ref{sec:datasets}) were consistently less important (Figure~\ref{fig:confidences}). This indicated that the agent avoided exploring consecutive PPIs.

Previous research on KGs has shown that node degree distribution, the number of adjacent edges for each KG node, can significantly bias predictions \citep{degree_bias_ev1, degree_bias_ev2}. Specifically, \textit{inspection bias}, a type of degree bias, occurs when the KG is not uniformly inspected or sampled \citep{degree_bias_xswap}. Since PPIs are the most common relation type in \textit{MoA-net} (90\% of edges) (Table~\ref{tab:kg_dist}), protein nodes have a higher degree distribution than other node types. From this evidence, we hypothesized that the agent was circumventing denser parts of the KG, creating an inspection bias. Although rule-based predictions merit a larger reward, the MARS agent exploits associative patterns for a more reliable reward. To confirm the existence of degree bias, we tested MARS upon \textit{MoA-net-permuted}. As explained in Section~\ref{sec:datasets}, \textit{MoA-net-permuted} is a variant of \textit{MoA-net} in which edges are swapped while preserving node degree distribution. This tests the extent to which node degree drives predictions. Indeed, the lack of change amongst standard performance metrics suggested that node degree was largely responsible for predictions (Figure~\ref{fig:metrics}-\(\beta\)). Put simply, the agent gets lost when exploring the PPIs, so it avoids them.

\subsection{Identifying and mitigating degree bias improves performance}\label{subsec:id_mitigate_bias}

 To temporarily prohibit the models from using associative patterns as in Section \ref{subsec:GBA}, we \textit{removed} inverse edges from \textit{MoA-net} and the corresponding metapath-based rules. Consequently, the performance metrics were poor, (\textit{e.g.,} standard MRR consistently \(< 0.1\) (Figure~\ref{fig:metrics}-\(\gamma\))). This confirmed that the models relied on associative patterns for predictions.

Next, to confirm that the agent was getting lost within the PPIs, we employed MARS' automatic trimming step. As explained in Section~\ref{sec:datasets}, \textit{MoA-net-10k} is a variant of \textit{MoA-net} with fewer PPIs. We tested each of \(\text{MARS}_{P_{2H}}\), \(\text{MARS}_{\text{naive}}\), and PoLo with the same parameters upon on \textit{MoA-net-10k} (Figure~\ref{fig:metrics}-\(\gamma\)). As before, inverse edges were excluded. Since trajectory length was set to \(L=4\), MARS' trimming approach also automatically removed drug-BP triples from the validation/test sets that were no longer connected via directed paths of length \(\leq 4\), resulting in 100 and 90 triples, respectively. Metrics were markedly improved for PoLo, \(\text{MARS}_{\text{naive}}\), and \textit{particularly} for \(\text{MARS}_{P_{2H}}\), in comparison to the full \textit{MoA-net} without inverse edges (Figure~\ref{fig:metrics}-\(\alpha\)). To ensure this improvement was not simply the result of a reduced test set, we tested each of the three approaches upon \textit{MoA-net} with 100 sampled test triples, which showed no change (see Appendix~\ref{subsec:ablation}).

While removing inverse edges improved metrics, a shortcut-aware system should achieve high \textit{performance} even with the shortcut present \citep{marconato2024bears}. We addressed this next.

\subsection{\(\text{MARS}_{P_{2H}}\) retains performance amongst rule-based predictions}

To restore MARS' option to use reasoning shortcuts, we re-introduced inverse edges to \textit{MoA-net-10k}. Thereafter, we tested each of \(\text{MARS}_{P_{2H}}\), \(\text{MARS}_{\text{naive}}\), and PoLo again (Figure~\ref{fig:metrics}-\(\epsilon\)). While each approach was optimized for \textit{standard} MRR, pruned metrics indicated how well positive predictions aligned with rules. In Figure~\ref{fig:metrics}-\(\epsilon\), one can see that both MARS variants and PoLo achieved standard metrics on par with or better than \textit{MoA-net-10k} without inverse edges (Figure~\ref{fig:metrics}-\(\gamma\)). However, \(\text{MARS}_{P_{2H}}\) also achieved pruned metrics comparable to its standard metrics, showing improved \textit{calibration} relative to PoLo and \(\text{MARS}_{\text{naive}}\). Finally, as in Section \ref{subsec:p2h_reveals_bias}, XSwap was used on \textit{MoA-net-10k} to assess the susceptibility of \(\text{MARS}_{P_{2H}}\) to degree bias. Unlike in Section~\ref{subsec:p2h_reveals_bias}, there was now no evidence for degree bias (see Appendix~\ref{app:final_xswap}).

\subsection{External validation of \(\text{MARS}_{P_{2H}}\) on \textit{MoA-net-10k}}

\highlight{In comparison to baseline methods, MARS' \textit{standard} metrics were on par with that of \mbox{\(\text{MARS}_{naive}\)} and PoLo, and sometimes outperformed by that of MINERVA (\mbox{Table~\ref{tab:benchmarks}}). However, \mbox{\(\text{MARS}_{P_{2H}}\)'s} \textit{pruned} metrics outperformed both standard and pruned metrics for all baseline models. Since \mbox{\(\text{MARS}_{P_{2H}}\)'s} \textit{pruned} metric intervals overlapped with reported intervals for MINERVA, PoLo, and \mbox{\(\text{MARS}_{naive}\)}, significance was verified based on Mann-Whitney U tests \mbox{\citep{mannwhitneyU}} (\mbox{\(\alpha=0.05\)}, one-sided) with Holm-Bonferroni correction for multiple testing \mbox{\citep{bonferroni}}, and \mbox{\(p\)-values} are reported in \mbox{Appendix~\ref{app:pvalues}}.}

Since MINERVA does not, by design, utilize rules for guidance, it suffers the same reasoning shortcuts as PoLo and \(\text{MARS}_{naive}\). \highlight{This is evident by its low pruned metrics reported in \mbox{Table~\ref{tab:benchmarks}}.} In contrast to MINERVA, DWPC suffers the opposite limitation: predictions are  based \textit{only} on metapath-based rules. \(\text{MARS}_{P_{2H}}\)'s \textit{pruned} metrics, which are directly comparable, also outperform DWPC.

Finally, as mentioned in Section~\ref{sec:datasets}, several drug-BP pairs corresponding to known MoAs in \textit{DrugMechDB} were included in the \textit{MoA-net} test set. Of these, 33 pairs remained within \textit{MoA-net-10k}'s test set. Based on the sequence of drugs, proteins, and BPs in an MoA, \(\text{MARS}_{P_{2H}}\) recovered the correct MoA for all 33 pairs. Notably, \mbox{\(\text{MARS}_{P_{2H}}\)} correctly predicted MoAs which consisted of three and four hops, including that of norethisterone (\mbox{Eq.~\ref{norethisterone}}) to treat heavy menstrual bleeding, and dexamethasone acetate (\mbox{Eq.~\ref{dex_acetate}}) to treat a cancer known as Multiple myeloma \mbox{\citep{drugmechdb}}. \highlight{In contrast, MINERVA failed to connect the drugs and BPs for these longer-range MoAs.} Thus, this comprehensive benchmark highlights MARS' ability to achieve near state-of-the-art performance by effectively balancing domain-specific knowledge with the capacity to generalize beyond it.

\begin{table}[ht]
\caption{Performance of MARS upon \textit{MoA-net-10k} against baseline models. Metrics are presented as \(\textit{average}\pm\textit{standard deviation}\) across five independent training/testing iterations for all but DWPC, which is deterministic. The best of each standard (top) and pruned (bottom) metric, if applicable, are in bold. The second best metrics are italicized.}
\label{tab:benchmarks}
\centering
\small
\begin{tabularx}{\textwidth}{>{\raggedright\arraybackslash}X *{4}{>{\centering\arraybackslash}X}}
\toprule
\textbf{Model} & \textbf{Hits@1} & \textbf{Hits@3} & \textbf{Hits@10} & \textbf{MRR} \\
\midrule

\multicolumn{5}{l}{\textbf{Standard Metrics}} \\
\cmidrule(lr){1-5}

CompGCN  & 0.093\(\pm\)0.010 & 0.212\(\pm\)0.031 & 0.428\(\pm\)0.043 & 0.201\(\pm\)0.011 \\

ComplEx  & 0.137\(\pm\)0.038 & 0.303\(\pm\)0.040 & 0.579\(\pm\)0.013 & 0.269\(\pm\)0.034 \\

MuRE  & 0.114\(\pm\)0.050 & 0.258\(\pm\)0.059 & 0.598\(\pm\)0.035 & 0.253\(\pm\)0.043 \\

PairRE  & 0.131\(\pm\)0.023 & 0.296\(\pm\)0.035 & 0.601\(\pm\)0.028 &  0.271\(\pm\)0.022 \\

RotatE  & 0.123\(\pm\)0.026 & 0.126\(\pm\)0.030 & 0.560\(\pm\)0.039 & 0.249\(\pm\)0.022 \\

 MINERVA  &  \textbf{0.342\(\pm\)0.016}     &\textbf{0.516\(\pm\) 0.042}    & 0.660\(\pm\)0.066    & \textbf{0.450\(\pm\)0.026} \\

 PoLo  & 0.272\(\pm\)0.041 &0.462\(\pm\)0.054
&0.606\(\pm\)0.061 &  0.387\(\pm\)0.044  \\

\(\text{MARS}_{\text{naive}}\)  & \textit{0.330\(\pm\)0.031}       &0.482\(\pm\)0.066 &\textit{0.664\(\pm\)0.036}    &     \textit{0.433\(\pm\)0.036}    \\

 \(\text{MARS}_{P_{2H}}\)  &  0.230\(\pm\)0.007         &\textit{0.492\(\pm\)0.027}  & \textbf{0.684\(\pm\)0.03}    &  0.395\(\pm\)0.016           \\

\midrule
\multicolumn{5}{l}{\textbf{Pruned Metrics}} \\
\cmidrule(lr){1-5}

 Metapaths with DWPC  & 0.370 & 0.560 & 0.780 & 0.508 \\

 \highlight{MINERVA} & 0.130\(\pm\)0.019& 0.144\(\pm\)0.024&0.144\(\pm\)0.024& 0.137\(\pm\)0.021\\

  PoLo & 0.170\(\pm\)0.049& 0.228\(\pm\)0.061&0.228\(\pm\)0.061& 0.198\(\pm\)0.052\\

  \(\text{MARS}_{\text{naive}}\) & \textit{0.220\(\pm\)0.049}&   \textit{0.238\(\pm\)0.048}&\textit{0.238\(\pm\)0.048}&\textit{0.229\(\pm\)0.048}\\

 \(\text{MARS}_{P_{2H}}\)  &  \textbf{0.394\(\pm\)0.026}&\textbf{0.644\(\pm\)0.034}&\textbf{0.788\(\pm\)0.018}&\textbf{0.535\(\pm\)0.02}\\
\bottomrule
\end{tabularx}
\end{table}

\begin{flalign}
\textsl{Norethisterone} &\xrightarrow{upregulates} \textsl{Progesterone receptor} \notag \\
&\xrightarrow{interacts} \textsl{Gonadotropin releasing hormone} \notag \\
&\xrightarrow{participates} \textsl{Luteinizing hormone secretion} &
\label{norethisterone}
\end{flalign}

\begin{flalign}
\textsl{Dexamethasone Acetate} &\xrightarrow{upregulates} \textsl{GC receptor} \xrightarrow{interacts} \textsl{Annexin A1} \notag \\
&\xrightarrow{interacts} \textsl{Phospholipase} \xrightarrow{participates} \textsl{Leukotriene biosynthesis} &
\label{dex_acetate}
\end{flalign}

\section{Discussion}\label{sec12}

NeSy approaches are sometimes portrayed as more trustworthy than their black-box counterparts, partially due to increased interpretability \citep{trustworthy, lauren_neuroai}. Here, we presented a NeSy RL approach, \(\text{MARS}_{P_{2H}}\), which promotes interpretability by deconvoluting drug MoAs. Specifically, through our novel algorithm, two-hop joint probabilities (\(P_{2H}\)), MARS learned weights corresponding to rules representing MoA patterns; each weight served as a proxy for each rule's importance. However, these insights revealed a new issue: NeSy RL approaches on KGs are susceptible to reasoning shortcuts. Specifically, in our study, predictions were driven by node degree bias. Ultimately, MARS’ interpretability called the trustworthiness of such approaches to question.

To address this, we considered Marconato \textit{et al.}'s \citep{marconato2024bears} desiderata for a shortcut-aware NeSy system. Specifically, on \textit{MoA-net-10k}, \(\text{MARS}_{P_{2H}}\) showed both competitive \textit{performance} as well as \textit{calibration} in comparison to other models. Notably, however, measuring \textit{calibration} is challenging in this domain. While rule-based predictions, measured through pruned metrics, follow the expected semantics for MoA deconvolution, we can not determine whether every \textit{other} prediction follows \textit{unintended} semantics. For example, in the classic MNIST addition task introduced in Section~\ref{sec:intro}, the misclassification of a handwritten `2' as `3' and vice versa would still amount to the same sum. Thus, reasoning shortcuts can be objectively identified. On the contrary, while we provide evidence that predictions using associative patterns are \textit{largely} affected by node degree bias, we can not determine whether such patterns \textit{always} reflect a reasoning shortcut. Finally, regarding \textit{cost effectiveness}, \(\text{MARS}_{P_{2H}}\) can be applied to any KG, serving as a generalizable mitigation strategy.

\subsection{Limitations and prospective directions}

MARS also has several limitations. Firstly, we note that the \textit{cost effectiveness} achieved by \(\text{MARS}_{P_{2H}}\) was limited to \textit{MoA-net-10k}, a trimmed version of \textit{MoA-net}. While we automated this trimming step, such a strategy does not make use of all available information. To scale \(\text{MARS}_{P_{2H}}\) to denser KGs and maintain its shortcut-aware status, several future directions could be explored. For instance, one could merge similar, high-degree nodes or rely upon domain knowledge, like the identification of promiscuous proteins \citep{promiscuous_proteins}, to make more informed choices about edge trimming or masking. \highlight{Notably, a recent study \mbox{\citep{pesquita}} developed a NeSy RL method, REx, which attenuates structural KG biases by incorporating information content (IC), which encodes information about node degree and type, directly into the reward function. The integration of IC into MARS' reward function could serve as an alternative mitigation strategy for node degree bias. Importantly, it could, potentially, reduce the need for KG pruning beforehand, thereby improving the generalizability of MARS to KGs which are denser or have imbalanced node and edge type distributions.

Additionally, as mentioned in \mbox{Section~\ref{p2h_updates}}, the \mbox{\(P_{2H}\)} updates assume independence between fragments. While this assumption was made to reduce computational complexity, it does not necessarily reflect the reality of the biology, in which such molecular interactions are not typically independent. Therefore, rewarding partial metapath matches without such an assumption is a key future direction. In addition to addressing methodological limitations, prospective studies could explore biological improvements. For example, as the DrugMechDB \mbox{\citep{drugmechdb}} continues to curate MoAs, future studies could utilize more ground-truth MoAs than the 33 used here. This could also, potentially, allow the incorporation of more complex MoAs. Alternatively, other future directions might involve the inclusion of binding or expression values, or more specific protein subclasses, which could help to better understand the patterns found here. For instance, the weights reported in \mbox{Figure~\ref{fig:confidences}} show that paths involving a \mbox{\(\texttt{downregulates}\)} relation were often weighted higher than paths of equal length with an \mbox{\(\texttt{upregulates}\)} relation. Including such information could help one understand whether this occurs due to a structural bias in the KG or a potential biological explanation.}

In summary, our study highlights a key concern in which the behavior of some NeSy RL approaches could be attributed to node degree bias, rather than meaningful, domain-specific concepts. The interpretability of our approach, \(\text{MARS}_{P_{2H}}\), allowed insight into this reasoning shortcut. Therefore, we question whether such shortcuts are identifiable amongst black-box approaches. Additionally, by testing a NeSy approach upon a novel applied task, MoA deconvolution, we could flag down patterns, like associative ones, which were plausible yet arguably less meaningful to biomedical researchers. Therefore, our study emphasizes the importance of testing interpretable models, like NeSy ones, in an applied domain. Finally, while our study honors the desiderata for shortcut-aware NeSy systems, we also examined the extent to which they were applicable to a biomedical domain.

\section{Conclusion}\label{sec13}

We propose a novel prediction task for NeSy approaches on biomedical KGs: mechanism-of-action (MoA) deconvolution. In contrast to previous DD approaches, MoA deconvolution utilizes model interpretability to uncover the molecular mechanisms behind medicinal drugs. We also constructed a publicly available KG, \textit{MoA-net}, for evaluating this task. To predict drug MoAs alongside indications, we designed the MoA Retrieval System (MARS). Relative to previous NeSy approaches, MARS has enhanced interpretability as it dynamically learns weights corresponding to logical rules. We showed that, with respect to the three desiderata for reasoning-aware NeSy systems, MARS has improved \textit{calibration} and \textit{cost effectiveness} compared to its predecessors, thereby enabling the identification \textit{and} mitigation of a reasoning shortcut based on node degree bias. \highlight{However, since shortcut mitigation was achieved upon a pruned, task-specific KG, a key future direction includes the generalization of MARS upon larger, denser KGs.}

\newpage

\backmatter

\section*{Declarations}

\subsection{Ethics approval and consent to participate}

Not applicable.

\subsection{Consent for publication}

Not applicable.

\subsection{Availability of data and materials}

The \textit{MoA-net} KG used in this study is publicly available in the MoA-net repository: \url{https://github.com/laurendelong21/MoA-Net}\citep{MoA_net_code}.

\noindent MARS code is publicly available in the MARS repository: \url{https://github.com/laurendelong21/MARS}\citep{MARS_code}.

\subsection{Competing interests}

YG and DDF were employees of Enveda during the course of this work and have real or potential interest in the company. LND was a temporary employee of Lyzeum Ltd but began employment after the course of this work. PG and LND are current employees of Cancer Research UK (CRUK), but CRUK is not affiliated or associated with this work. All other authors declare no competing interests.

\subsection{Funding}

During completion of this study LND was funded by the University of Edinburgh Informatics Graduate School through the Global Informatics Scholarship as well as the Alan Turing Institute Enrichment Scheme. LND partially conducted this work during an internship at Enveda. JDF and PG were funded by the National Institute for Health Research (NIHR) Artificial Intelligence and Multimorbidity: Clustering in Individuals, Space and Clinical Context (AIM-CISC) grant NIHR202639. The views expressed are those of the author(s) and not necessarily those of the NIHR or the Department of Health and Social Care.

\subsection{Authors' contributions}

LND designed the methodology, wrote the software, and conducted the investigation. YG and LND did data curation and visualization. DDF and LND conceptualized the project. DDF, JDF, and PG supervised the project. All authors took part in writing the manuscript.

\subsection{Acknowledgements}

 We thank our funding institutions for their support. We also thank Emile van Krieken and Guillermo Romero Moreno for their thoughtful feedback.

\begin{appendices}

\section{Metapaths}\label{app:metapaths}

As mentioned in Section~\ref{subsec:metapaths}, all metapaths involving mechanistic patterns were used:

\begin{table}[h!]
  \caption{Metapaths representing MoAs. Drugs are represented with a \(\textsl{Drug}\), proteins with a \(\textsl{Protein}\), and biological processes with \(\textsl{BP}\).}
  \small
\begin{tabular}{l}
    \hline \\
    \textbf{(1)} \(\texttt{downregulates}(\textsl{Drug}, \textsl{Protein}) \rightarrow \texttt{participates}(\textsl{Protein}, \textsl{BP}) \)\\
    
    \textbf{(2)} \(\texttt{upregulates}(\textsl{Drug}, \textsl{Protein}) \rightarrow \texttt{participates}(\textsl{Protein}, \textsl{BP})\)\\

    \textbf{(3)} \(\texttt{downregulates}(\textsl{Drug}, \textsl{Protein}) \rightarrow\) \\
    \(\hphantom{\texttt{downregulates}(\textsl{Drug}, \textsl{Protein})} \texttt{interacts}(\textsl{Protein}, \textsl{Protein}) \rightarrow \texttt{participates}(\textsl{Protein}, \textsl{BP})\)\\

    \textbf{(4)} \(\texttt{upregulates}(\textsl{Drug}, \textsl{Protein})\)\\
    \(\hphantom{\texttt{downregulates}(\textsl{Drug}, \textsl{Protein})} \rightarrow \texttt{interacts}(\textsl{Protein}, \textsl{Protein}) \rightarrow \texttt{participates}(\textsl{Protein}, \textsl{BP})\)\\

    \textbf{(5)} \(\texttt{downregulates}(\textsl{Drug}, \textsl{Protein}) \rightarrow \texttt{interacts}(\textsl{Protein}, \textsl{Protein}) \rightarrow\) \\
    \(\hphantom{\texttt{downregulates}(\textsl{Drug}, \textsl{Protein})} \texttt{interacts}(\textsl{Protein}, \textsl{Protein}) \rightarrow \texttt{participates}(\textsl{Protein}, \textsl{BP})\)\\

    \textbf{(6)} \(\texttt{upregulates}(\textsl{Drug}, \textsl{Protein}) \rightarrow \texttt{interacts}(\textsl{Protein}, \textsl{Protein}) \rightarrow\) \\
    \(\hphantom{\texttt{downregulates}(\textsl{Drug}, \textsl{Protein})}\texttt{interacts}(\textsl{Protein}, \textsl{Protein}) \rightarrow \texttt{participates}(\textsl{Protein}, \textsl{BP})\)\\
    \\
    \hline
    \\
\end{tabular}
\label{tab:metapaths}
\end{table}

\newpage

\section{Implementation}\label{app:implementation}

\begin{figure}[ht]
\begin{center}
\includegraphics[width=0.7\linewidth]{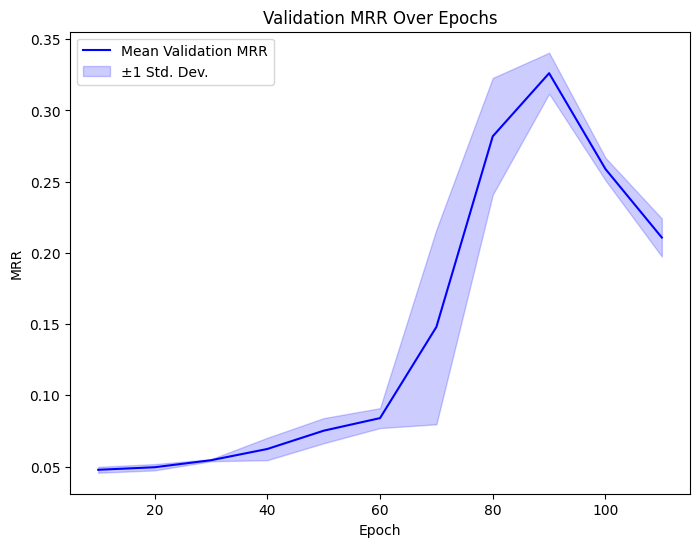}
\end{center}
\caption{Validation MRR over training epochs.}
\label{fig:val_mrr}
\end{figure}

MARS was implemented using TensorFlow (v. 2.10) and packaged in Python\footnote{https://github.com/laurendelong21/MARS}. The neural network structure underlying the LSTM is implemented as in Liu \textit{et al.} \citep{liu2021neural}, which is also drawn from MINERVA \citep{minerva}. The Adam optimizer \citep{KingBa15} was used with REINFORCE \citep{williams1992simple} to maximize rewards. Hyperparameter optimization was done via grid search \citep{feurer2019hyperparameter}; further details are within Appendix~\ref{app:hp_selection}. MARS was trained to optimize MRR, with early stopping determined by validation MRR (Fig.~\ref{fig:val_mrr}). For training MARS, one A40 Nvidia GPU \citep{nvidia_a40} and two AMD EPYC Milan 7413 CPU nodes \citep{amd_epyc_7413} were used.
 
\section{Hyperparameter selection}\label{app:hp_selection}

Here, hyperparameter selection is described. The hyperparameter search space was high-dimensional - there were fifteen hyperparameters and a continuous range of possibilities for several of them. Therefore, some hyperparameter values were chosen based upon the needs of this study and previous work \citep{liu2021neural}. Table~\ref{hyperparameters_fixed} describes the hyperparameters which were fixed for every model. The other hyperparameters were optimized via grid search \citep{feurer2019hyperparameter} ona restricted set of values. Table~\ref{hyperparameters} describes the hyperparameter search space for optimization. Finally, table~\ref{hyperparameters_chosen} describes the best hyperparameters for the final results in Fig.~\ref{fig:metrics}-\(\epsilon\).

\begin{table}[ht!]
\caption{Fixed hyperparameter settings}
\label{hyperparameters_fixed}
\begin{tabular}{p{0.22\linewidth} p{0.50\linewidth} p{0.22\linewidth}}
{\bf Hyperparameter}  &{\bf Description}&{\bf Value}
\\ \hline \\
embedding size         & size of the relation and entity embeddings & 256 \\
LSTM layers            & number of LTSM layers & 2 \\
test rollouts & number of times each query (source-terminal node pair) is made or attempted during testing & 50\\
max branching & maximum number of outgoing edges per node shown to the agent in an episode & 150 \\
\(\gamma\) (gamma) & discount factor as implemented in REINFORCE \citep{williams1992simple} & 1\\
positive reward & reward for finding a true pair & 1\\
negative reward & penalty for failing to find a true pair & 0
\end{tabular}
\end{table}

\begin{table}[ht]
\caption{Hyperparameter search space for grid search optimization \citep{feurer2019hyperparameter}. Note that \(\lambda\) is not applicable for MINERVA, and \(\alpha\) is only applicable to MARS.}
\label{hyperparameters}
\begin{tabular}{p{0.30\linewidth} p{0.40\linewidth} p{0.24\linewidth}}
{\bf Hyperparameter}  &{\bf Description}&{\bf Search space}
\\ \hline \\
\(\lambda\) (Lambda)         & ratio at which the second summand, or reward, is applied relative to the first summand, or reward, in the reward function & \{5, 8, 10\} \\
\(\alpha\) (alpha)            & how dramatically weight updates should be made (if applicable) & \{0.001, 0.01, 0.1\} \\
learning rate             & learning rate of the optimizer & \{0.0001, 0.001, 0.01\} \\
hidden size & size of hidden layers & \{64, 128, 256\}\\
batch size & size of sampled mini-batch for training & \{128, 256\}\\
rollouts & number of times each query (source-terminal node pair) is made or attempted during training & \{50, 100\}\\
\(\gamma_{\text{baseline}}\) (gamma baseline) & discount factor for the baseline as implemented in MINERVA \citep{minerva} & \{0.05, 0.5\}\\
\(\beta\) (beta) & entropy regularization factor as implemented in MINERVA \citep{minerva} & \{0.025, 0.05\}
\end{tabular}
\end{table}

\begin{table}[ht]
\caption{Best hyperparameters from Table~\ref{hyperparameters} for the experiments in Fig.~\ref{fig:metrics}-\(\epsilon\).}
\label{hyperparameters_chosen}
\begin{tabular}{p{0.22\linewidth} p{0.22\linewidth} p{0.22\linewidth}p{0.22\linewidth}}
{\bf Hyperparameter}  &{\bf \(\text{MARS}_{P_{2H}}\)}&{\bf \(\text{MARS}_{\text{naive}}\)}&{\bf{PoLo}}
\\ \hline \\
\(\lambda\) & 10 & 5 & 5 \\
\(\alpha\) &  0.001 & 0.001 & - \\
learning rate & 0.0001 & 0.0001 &  0.0001 \\
hidden size & 256 & 256 & 64 \\
batch size & 128 & 256 & 256\\
rollouts & 100 & 100 & 50 \\
\(\gamma_{\text{baseline}}\) & 0.05 & 0.5 & 0.5 \\
\(\beta\) & 0.025 & 0.05 & 0.05 \\
\end{tabular}
\end{table}
\newpage

The baseline KGE models were trained using the PyKEEN framework (v1.10.1) \citep{ali2021pykeen}. KGE models were trained using PyKEEN’s hyperparameter optimization pipeline over 30 trials using as initial parameters the best configurations from \citep{rivas2022kgem}. Hyperparameters for each model were optimized on MRR over a drug-BP link prediction task for the previously-described splits. The DWPC baseline had only one hyperparameter: the damping exponent, which determines how much node degree affects predictions. Since a damping exponent of \(w = 0.4\) was found to be optimal on other biomedical prediction tasks \citep{himmelstein2015heterogeneous}, this value was also used in this study. Finally, network algorithms were implemented in NetworkX (v3.1) \citep{networkx}.

\section{PoLo metrics without associative rules}\label{app:polo}

 The results of Liu \textit{et al.} \citep{liu2021neural} with PoLo were reproduced on the \textit{Hetionet} KG \citep{hetionet} using the same parameters and data splits as reported by Liu \textit{et al.} In contrast to Liu \textit{et al.}, only directed metapaths of length \(L \leq 4\) were included as rule bodies (as in Appendix~\ref{subsec:metapaths}). These metapaths served as the metapath-based rules for PoLo. Notably, these metapaths excluded the associative metapath mentioned in section~\ref{subsec:GBA}:

\(\texttt{treats}(\textsl{Drug}_1, \textsl{Disease}) \Leftarrow \texttt{causes}(\textsl{Drug}_1, \textsl{Side Effect}) \land\\
\hphantom{\texttt{treats}(\textsl{Drug}_1, \textsl{Disease}) \Leftarrow } \texttt{causes}(\textsl{Drug}_2, \textsl{Side Effect}) \land \texttt{treats}(\textsl{Drug}_2, \textsl{Disease})\)

 Despite the most-used metapath-based rule being absent, PoLo achieved the same standard metrics as reported by Liu \textit{et al.} (Table~\ref{tab:polo_metrics}).

\begin{table}[htbp]
\caption{Performance evaluations of PoLo upon \textit{Hetionet} as reported in Liu \textit{et al.} \citep{liu2021neural} (\textit{average} across five independent training/testing iterations) and PoLo upon \textit{Hetionet} \textit{without} associative rules (\textit{average}\(\pm\)\textit{standard deviation} across four independent training/testing iterations.)}
\label{tab:polo_metrics}
\small
\begin{tabular}{lllll}
\hline
\textbf{rule types} & \textbf{Hits@1} & \textbf{Hits@3} & \textbf{Hits@10} & \textbf{MRR} \\
\hline

 associative (Liu \textit{et al.} \citep{liu2021neural}) & 0.314 & 0.428
&0.609 &  0.402  \\

 mechanistic (this study) &  0.328\(\pm\)0.046 &
0.465\(\pm\)0.037 & 0.656\(\pm\)0.044 & 0.431\(\pm\)0.035   \\

\end{tabular}
\end{table}

\newpage

\section{Ablation study}\label{subsec:ablation}

\begin{figure}[hbt]
    \centering
    {\includegraphics[width=0.5\textwidth]{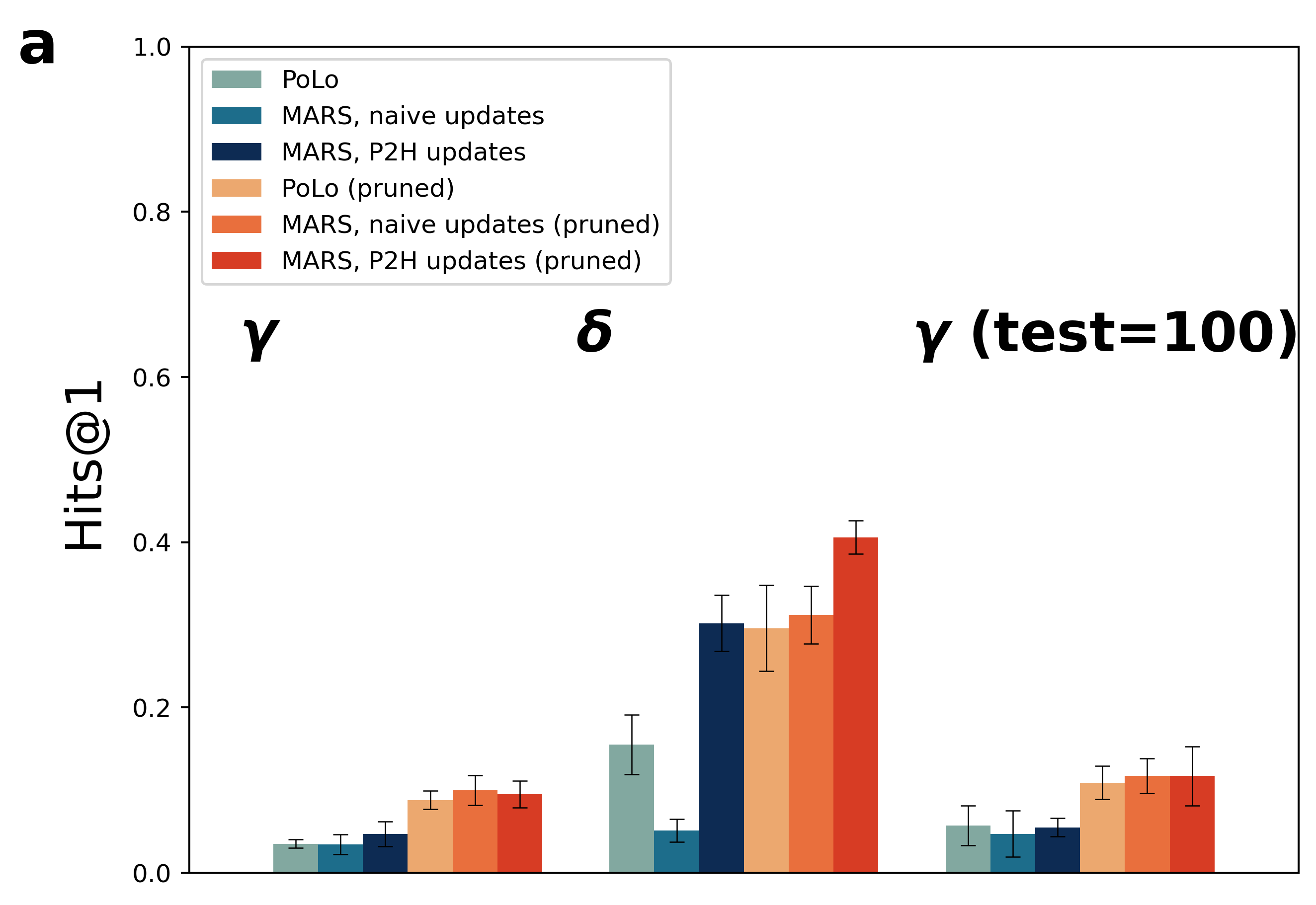}}{\includegraphics[width=0.5\textwidth]{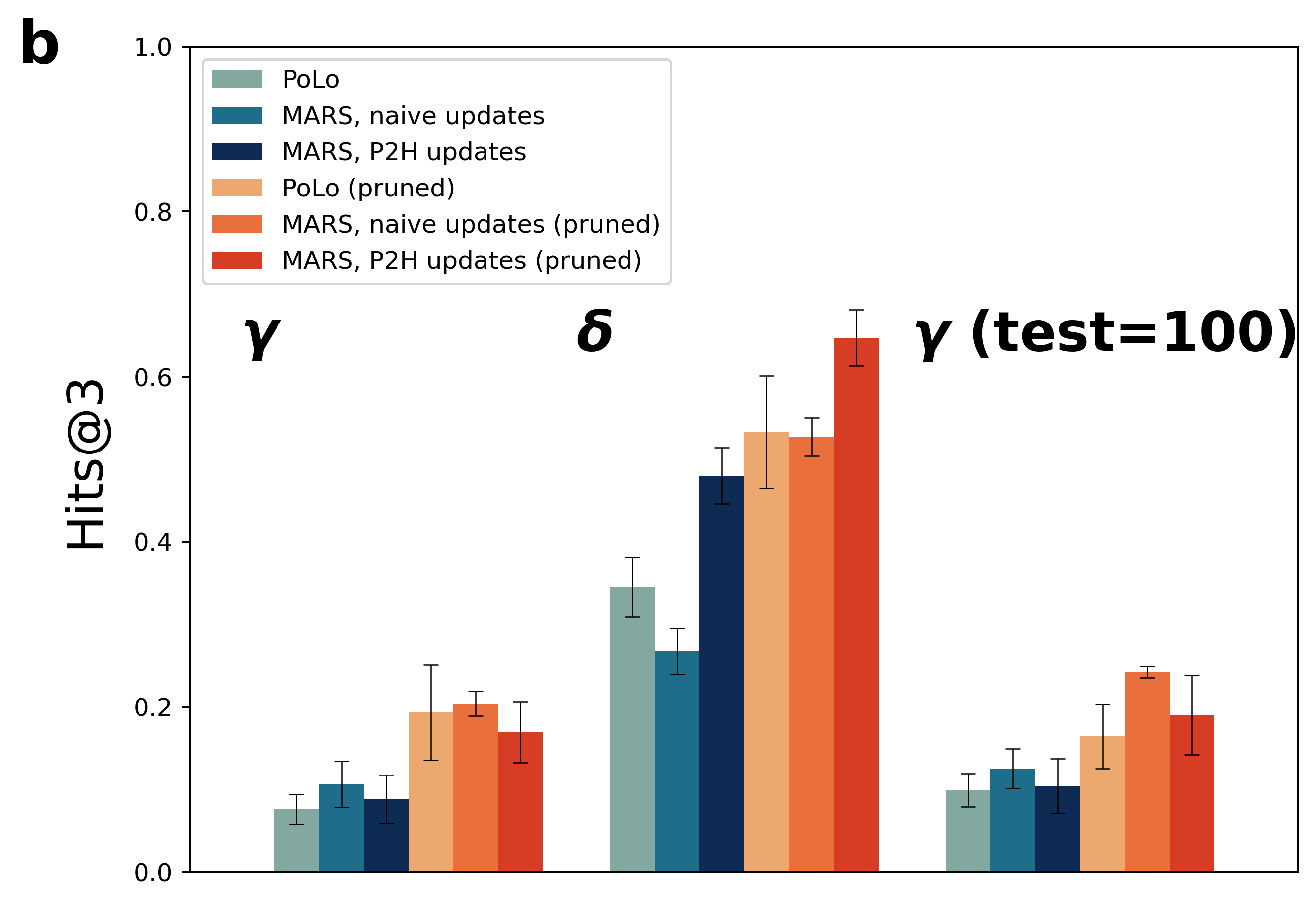}} \\{\includegraphics[width=0.5\textwidth]{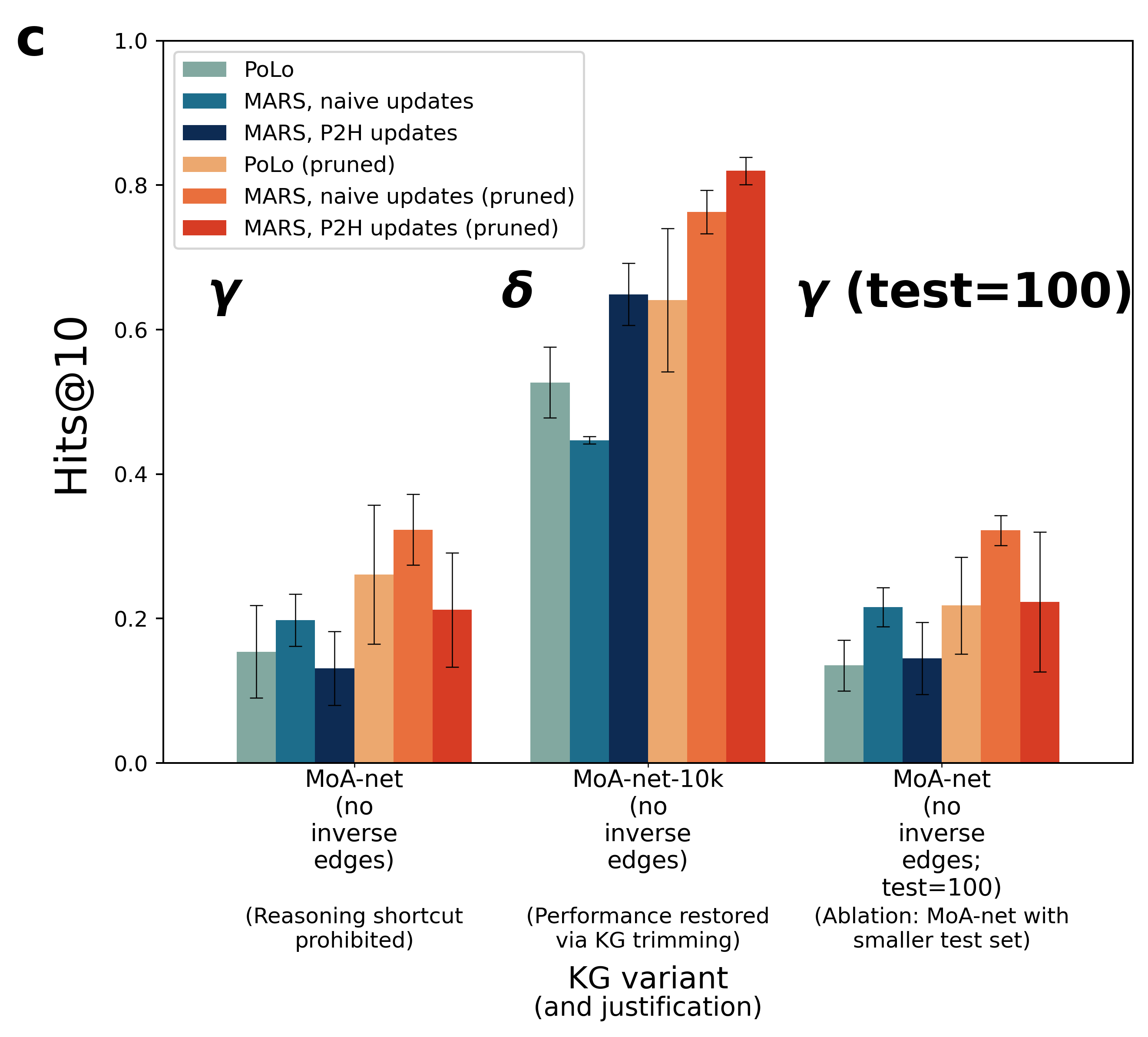}}{\includegraphics[width=0.5\textwidth]{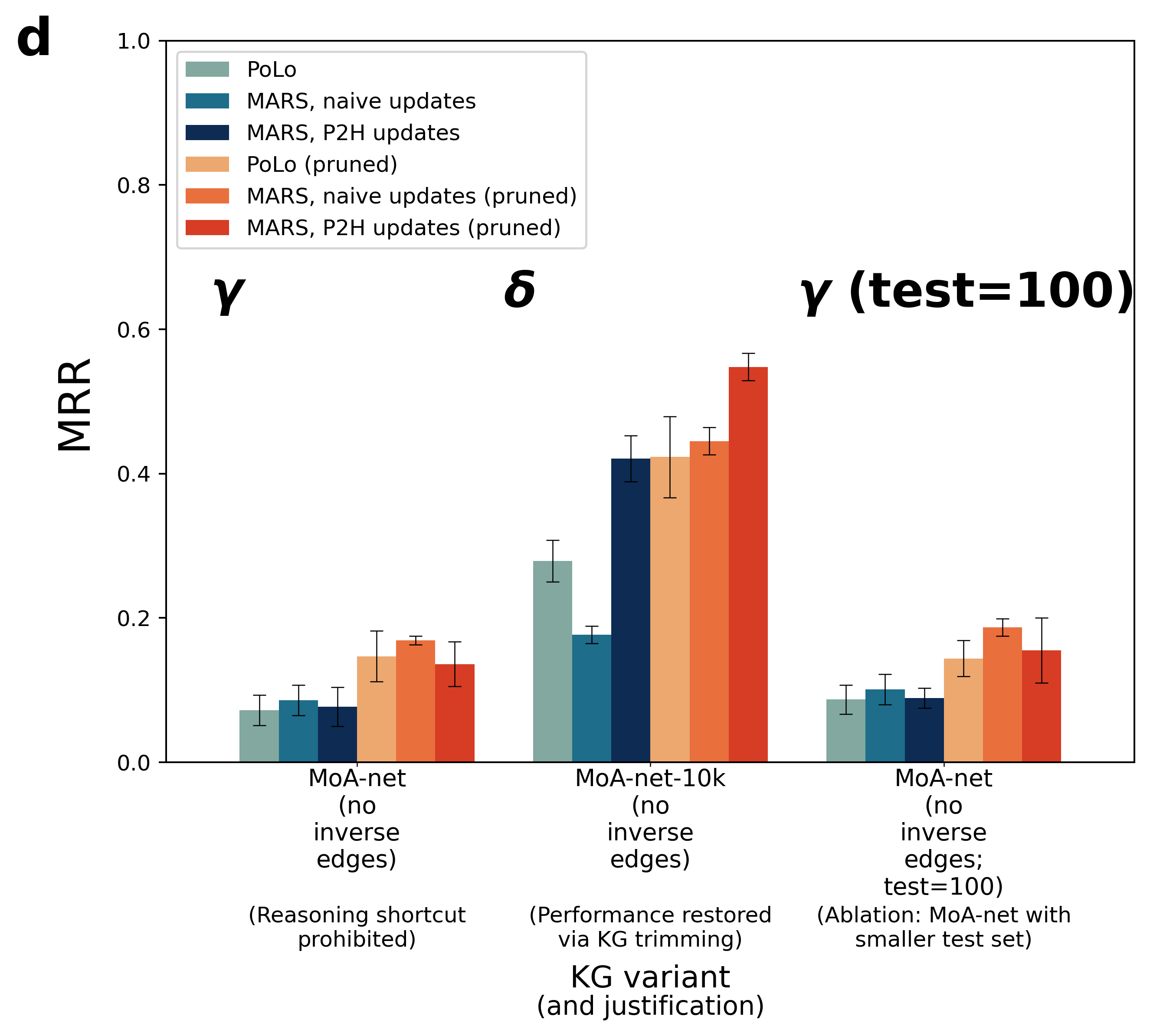}}
    \caption{Performance evaluations upon \textit{MoA-net} (no inverse edges) with a test set of 100 triples. Metrics are presented as the average, with error bars representing standard deviation across five independent training/testing iterations. The lack of change between \(\gamma\) and \(\gamma\) (test=100) indicates that a reduction in test set size is not responsible for improvements observed in \(\delta\).}\label{metrics-smaller-test}
\end{figure}

To validate that the results on \textit{MoA-net-10k} were not due to a reduced test set, the effects of reducing the test set size for \textit{MoA-net} (\(n_{test}\)=100) were assessed. The lack of change between Fig.~\ref{metrics-smaller-test}-\(\gamma\) and \(\gamma\) (test=100) indicates that a reduction in test set size is not responsible for improvements observed in Fig.~\ref{metrics-smaller-test}-\(\delta\).

\newpage

\section{XSwap permutations: \(\text{MARS}_{P_{2H}}\) on \textit{MoA-net-10k}}
\label{app:final_xswap}

\begin{figure}[hbt]
    {\includegraphics[width=0.5\textwidth]{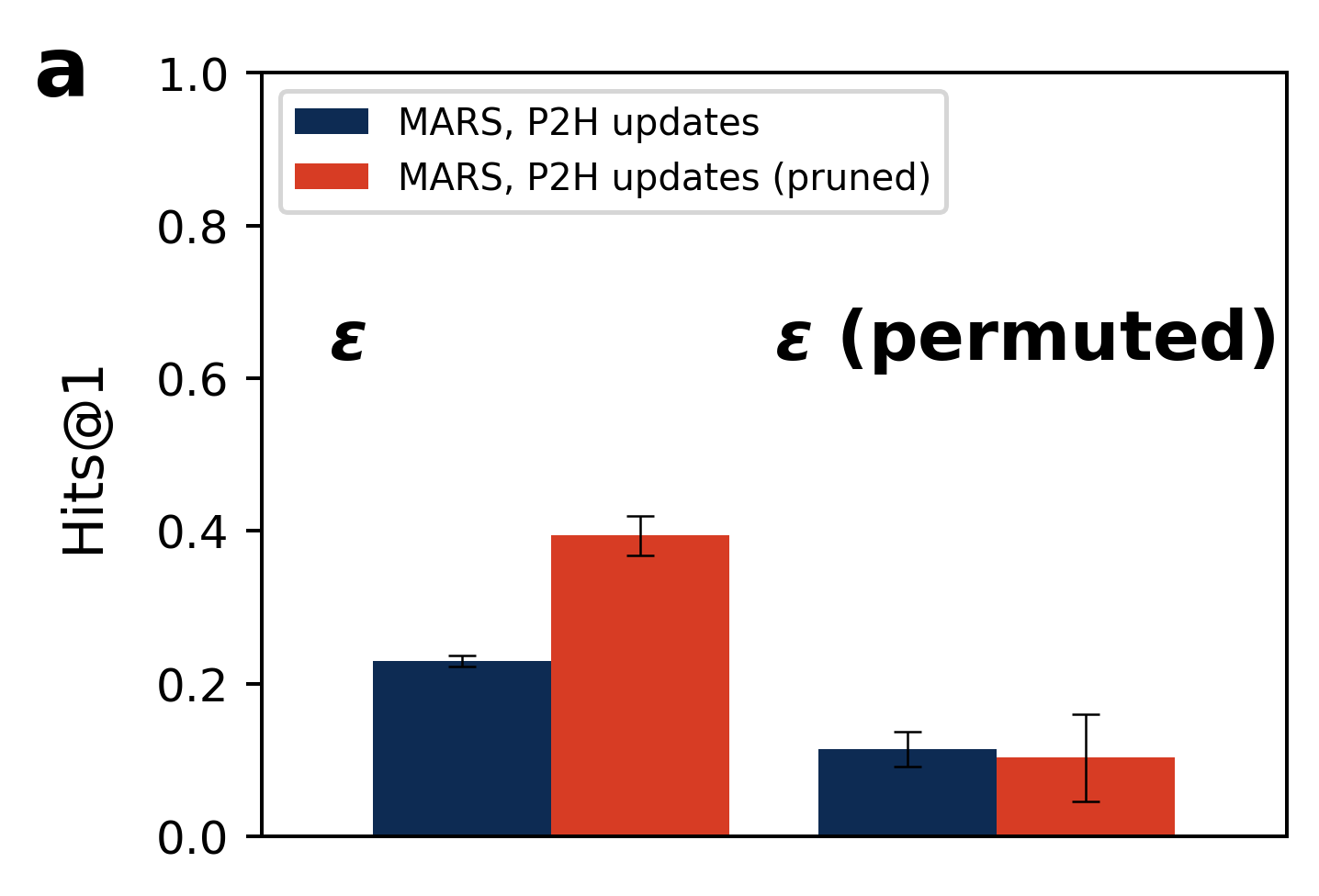}}{\includegraphics[width=0.5\textwidth]{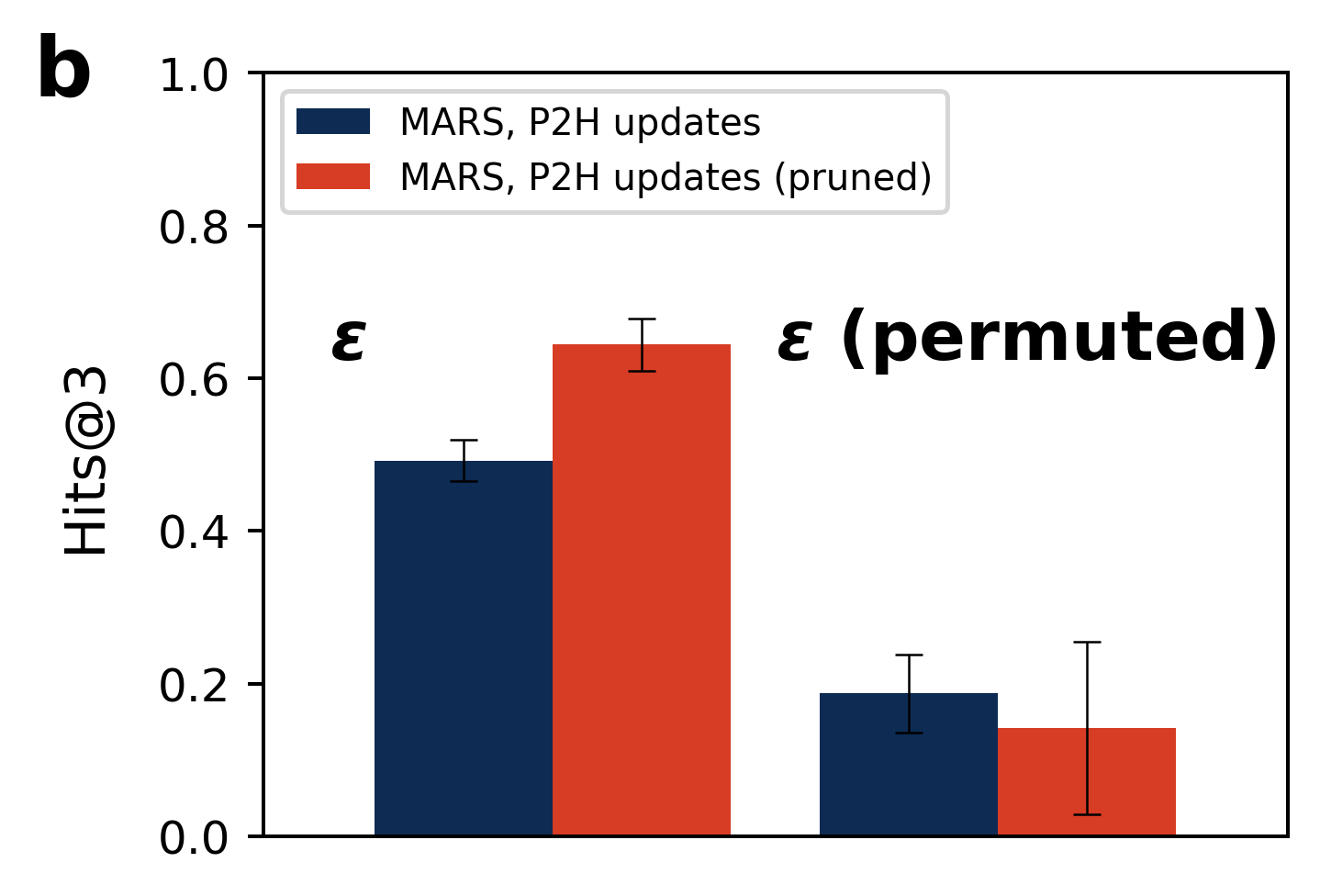}} \\{\includegraphics[width=0.5\textwidth]{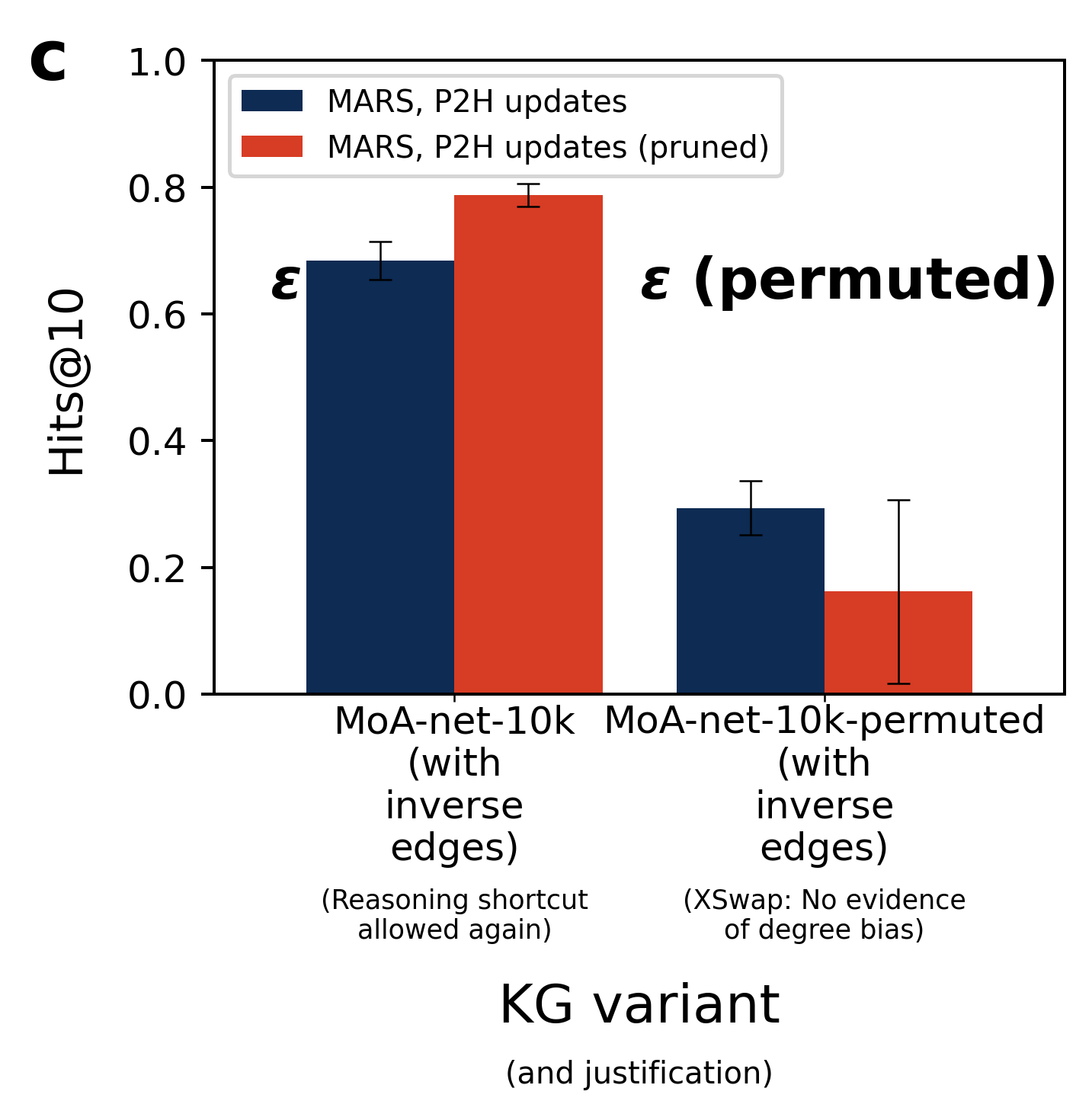}}{\includegraphics[width=0.5\textwidth]{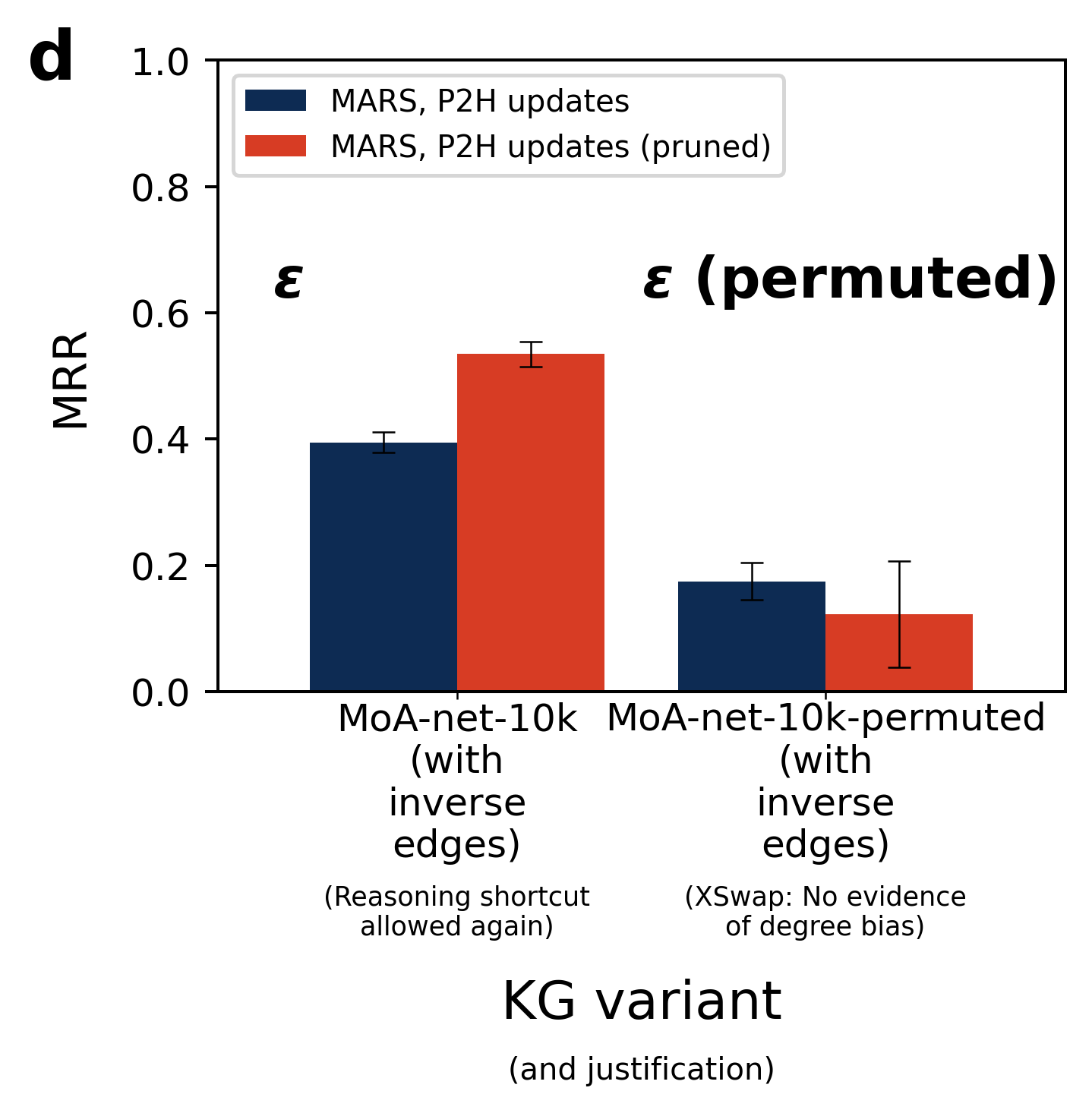}}
    \caption{Performance evaluations of \(\text{MARS}_{P_{2H}}\) on \textit{MoA-net-10k} as well as a permuted variant of \textit{MoA-net-10k} via the XSwap algorithm. Metrics are presented as the average, with error bars representing standard deviation across five independent training/testing iterations. A drop in performance metrics (\(\epsilon\) (permuted)) indicates that node degree was not the main driver in predictions made in \(\epsilon\).}\label{fig:final_xswap}
\end{figure}

By using the XSwap algorithm as in Section \ref{subsec:p2h_reveals_bias} as an ablation, it was possible to assess whether the prediction metrics achieved using \(\text{MARS}_{P_{2H}}\) on \textit{MoA-net-10k} were influenced by degree bias. This time, there was a stark decrease in performance metrics upon the permuted KG (Fig.~\ref{fig:final_xswap}). This showed that predictions made by \(\text{MARS}_{P_{2H}}\) were due to factors beyond node degree bias.

\section{Significance tests for reported metrics}\label{app:pvalues}

\begin{table}[ht]
\caption{Holm-Bonferroni adjusted, Mann-Whitney U test \(p\)-values between \(\text{MARS}_{P_{2H}}\) pruned metrics and all other MINERVA, PoLo, and MARS metrics upon \textit{MoA-net-10k}}
\label{tab:pvalues}
\centering
\small
\begin{tabularx}{\textwidth}{>{\raggedright\arraybackslash}X *{4}{>{\centering\arraybackslash}X}}
\toprule
\textbf{Model} & \textbf{Hits@1} & \textbf{Hits@3} & \textbf{Hits@10} & \textbf{MRR} \\
\midrule

\multicolumn{5}{l}{\textbf{Standard Metrics}} \\
\cmidrule(lr){1-5}

 MINERVA  &  0.021819     &0.021819    & 0.021325    & 0.015873 \\

 PoLo  & 0.021819 &0.021819
&0.021325 &  0.015873  \\

\(\text{MARS}_{\text{naive}}\)  & 0.021819       &0.021819 &0.021325    &     0.015873    \\

 \(\text{MARS}_{P_{2H}}\)  &  0.021819         &0.021819  & 0.021325    &  0.015873          \\

\midrule
\multicolumn{5}{l}{\textbf{Pruned Metrics}} \\
\cmidrule(lr){1-5}

 MINERVA & 0.017118& 0.017118&0.015993& 0.011925\\

  PoLo & 0.017118& 0.017118&0.015993& 0.011905\\

  \(\text{MARS}_{\text{naive}}\) & 0.017118&   0.017118&0.015993&0.011925\\

\bottomrule
\end{tabularx}
\end{table}

\end{appendices}


\bibliography{sn-bibliography.bib}

\end{document}